\documentclass[11pt]{article}

\usepackage[final]{acl}

\usepackage{times}
\usepackage{latexsym}
\usepackage[utf8]{inputenc}
\usepackage{csquotes}
\usepackage{graphicx}
\usepackage{multirow}
\usepackage{booktabs}
\usepackage{makecell}
\usepackage{amsmath} 

\usepackage[T1]{fontenc}

\usepackage[utf8]{inputenc}

\usepackage{microtype}

\usepackage{inconsolata}

\usepackage{graphicx}
\usepackage[most]{tcolorbox}
\usepackage{subcaption}

\usepackage{hyperref}

\title{CoPiT: Cognitive Pivot Translation for Digraphic Low-Resource Mongolian in the Traditional Script}

\author{
Burte Bayarsaikhan\thanks{Equal contribution.}\textsuperscript{1} \quad
Serynn Kim\footnotemark[1]\textsuperscript{2} \quad
Buru Chang\thanks{Corresponding author.}\textsuperscript{1} \\
\textsuperscript{1}Korea University \quad
\textsuperscript{2}Hankuk University of Foreign Studies \\
\texttt{\{burtebay,buru\_chang\}@korea.ac.kr}\\
\texttt{serynn@hufs.ac.kr}
}

\begin{document}
\maketitle


\newcommand{\copit}{\textit{{CoPiT}}}

\begin{abstract}\label{sec:0_abstract}
Low-resource languages remain challenging for machine translation, and Mongolian is a representative case. As a digraphic language, Mongolian is written in both Cyrillic and Traditional scripts, which exhibit a severe imbalance in data availability. While the Cyrillic script is relatively well-resourced, the Traditional script remains extremely data-scarce and orthographically ambiguous, leading to substantial performance degradation in direct translation. We propose \copit{}, a cognitively motivated pivot-based translation pipeline that exploits this internal resource hierarchy by routing translation through the Cyrillic script. The pipeline explicitly resolves script-induced ambiguity in the Traditional script before translation, enabling more stable and accurate meaning transfer. Across multiple backbone models and target languages, \copit{} generally improves over direct translation, with substantial absolute BLEU improvements and up to 1.5–1.6$\times$ gains in COMET. These gains allow strong open-source models to match or outperform GPT-4.1 under comparable evaluation settings.
Beyond inference-time improvements, \copit{} enables the construction of synthetic parallel data directly from Traditional-script text, mitigating data scarcity in realistic low-resource scenarios. We release a new multi-script parallel dataset covering Mongolian in both scripts alongside English, Korean, and Russian. All datasets and code are publicly available at \url{https://github.com/bbyba/CoPiT.git}.
\end{abstract}
\section{Introduction}\label{sec:1_introduction}
Large language models (LLMs) have evolved from primarily monolingual systems to multilingual systems through large-scale multilingual pre-training. Despite this progress, LLMs continue to perform poorly on low-resource languages (LRLs), largely due to limited training data and sparse linguistic resources. This limitation is particularly evident in machine translation (MT), which relies heavily on large-scale parallel corpora that are often unavailable for LRLs \citep{raja-vats-2025-parallel}.  
\begin{figure}[t]
    \centering
    \includegraphics[width=0.85\columnwidth]{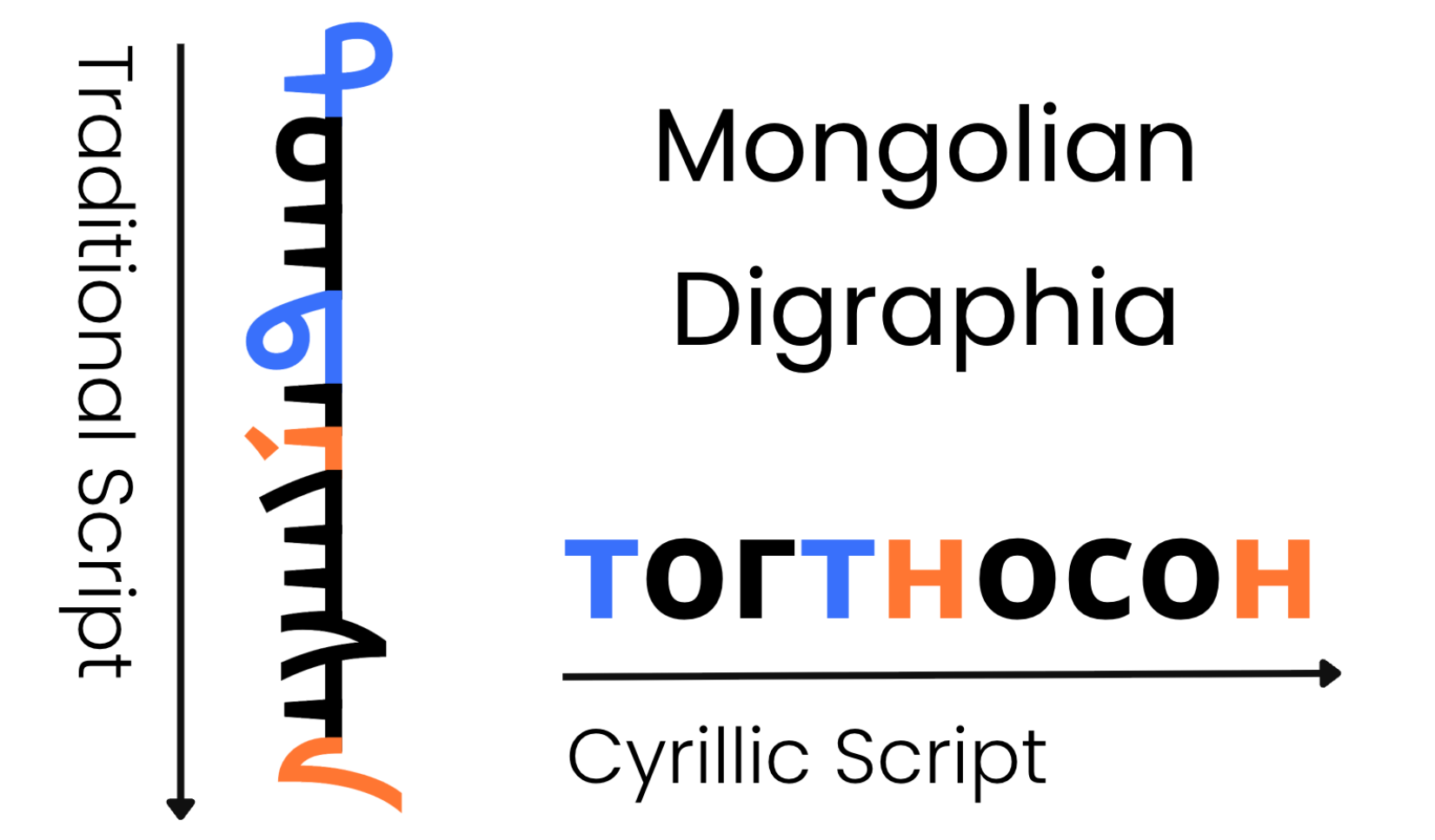}
    \caption{Mongolian digraphia. The same content can be written in Traditional script (left), with multiple surface forms and limited resources, and in better-resourced Cyrillic script (right).}
    \label{fig:1_scripts}
    \vspace{-1em}
\end{figure}

These challenges are particularly pronounced in Mongolian, a low-resource language with a unique \textbf{digraphic characteristic}, as it is written in both \textit{Cyrillic} and \textit{Traditional} scripts (see Figure~\ref{fig:1_scripts}). The Cyrillic script, which serves as the dominant modern form, is largely phonemic and exhibits a high degree of consistency in the sound representation. In contrast, the Traditional script, an archaic writing system, is written vertically, and its letter forms vary by positional context. Many phonological and morphological distinctions are not explicitly encoded in the script, such that a single written form may correspond to multiple plausible interpretations. This intrinsic ambiguity distinguishes the Traditional script structurally from its Cyrillic counterpart. For example, the Gemini-3-Pro-Preview API flags content written in the Traditional Mongolian script as harmful, as shown in Appendix~\ref{subsec:appendix_gemini}.

Beyond these structural differences, the two Mongolian scripts are associated with markedly different levels of linguistic and computational resources. The Cyrillic script has long served as the dominant writing system in modern Mongolian society following its institutional adoption through state language policies. As a result, the vast majority of textual data and language technologies are developed for the Cyrillic script. In contrast, the Traditional script, despite representing the same language, is reintroduced into official use through a policy-driven effort to promote traditional writing practices\footnote{The Law on the Mongolian Language, adopted on February 12, 2015, and amended with effect from January 1, 2025, mandates the use of both Cyrillic and Traditional scripts in official state and local government affairs; Official text available at \url{https://legalinfo.mn/mn/detail/10932}.}. Owing to this delayed revival, the Traditional script remains severely under-resourced, giving rise to a hierarchical resource imbalance within a single LRL. This imbalance leads current NLP systems to exhibit pronounced performance asymmetry between the two scripts, with poorer performance on the Traditional script.

To address this gap, we introduce a novel pipeline, \textit{\textbf{CoPiT} (\textbf{Co}gnitive \textbf{Pi}vot \textbf{T}ranslation}), for translating Mongolian written in the Traditional script into other languages. Our pipeline is motivated by how fluent Mongolian readers process the Traditional script in practice~\citep{altangerel2024classical}. Due to the dominance of the Cyrillic script in modern usage, many native speakers do not read the Traditional script in a purely direct manner. Instead, they implicitly map Traditional script forms to their Cyrillic counterparts and interpret them through this familiar representation, resolving script-dependent ambiguities in a phonologically informed way. Accordingly, we adopt Mongolian written in the Cyrillic script as an intermediate representation. While it remains low-resource, it offers a more phonemically transparent and comparatively better-resourced representation than the Traditional script, and thus aligns more effectively with both human reading strategies and the strengths of current LLMs.

Concretely, we decompose the MT task into two sequential stages: (1) conversion from Mongolian written in the Traditional script to the Cyrillic script, followed by (2) translation from the Cyrillic script to the target language. This design isolates the script-specific challenges of the Traditional script from downstream semantic translation while leveraging the relatively greater, though still limited, resources available for the Cyrillic script.

The Traditional-to-Cyrillic conversion is further factorized into three linguistically motivated steps that align with disambiguation processes used by fluent Mongolian readers: vowel harmony recovery, Latin-assisted normalization, and Cyrillic normalization. Collectively, these steps transform the underspecified Traditional script into a phonologically specified Cyrillic representation, thereby facilitating subsequent translation into the target language.
This explicit resolution of script-level ambiguity enables \copit{} to translate Mongolian written in the Traditional script without relying on large-scale parallel corpora. Instead, it requires only a small amount of data for learning script conversion.


Comprehensive experimental results demonstrate that our pipeline significantly improves translation accuracy across multiple target languages, including English, Korean, and Russian. When applied to GPT-4.1, zero-shot \copit{} increases BLEU-3 scores from 2.75 to 14.07 for English, and yields an average improvement from 2.0 to 12.05 across target languages compared to direct translation. In addition, fine-tuned \copit{} models on open-source backbones match or exceed GPT-4.1 performance under both reference-based and reference-free evaluations. Beyond this, \copit{} generates a synthetic parallel corpus of 8,034 sentence pairs aligning Traditional Mongolian with English, Korean, and Russian. This synthetic data proves useful beyond forward translation: it further improves forward translation quality via translation module fine-tuning and enables reverse-direction translation, demonstrating the broader utility of our approach under realistic low-resource constraints.
\begin{figure*}[t]
    \centering
    \includegraphics[width=\textwidth]{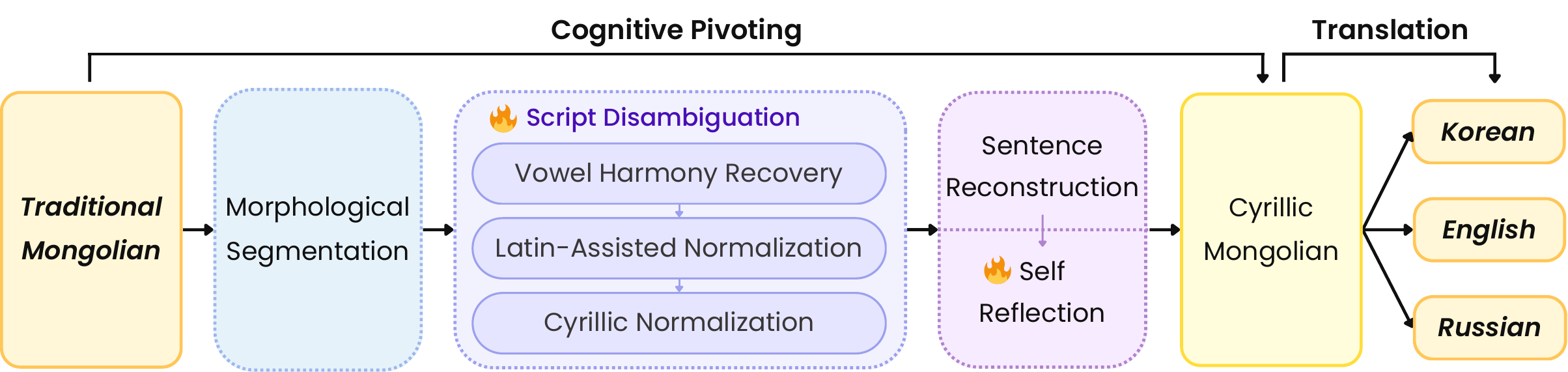}
    \caption{Overview of \copit, a pipeline translating Mongolian from the Traditional script via the Cyrillic script.}
    \label{fig:2_pipeline}
    \vspace*{-1em}
\end{figure*}


Our contributions are as follows:
\begin{itemize}
    \item We propose \copit{}, a cognitively motivated multi-stage pipeline that translates Traditional-script Mongolian by resolving script-induced ambiguity through Cyrillic pivoting.
    \item We release a multi-script parallel corpus aligning Traditional and Cyrillic Mongolian with English, Korean, and Russian, addressing a resource gap for digraphic low-resource MT.
    \item We show that \copit{} generally improves over direct translation across backbones and target languages, with fine-tuned open-source models matching or exceeding GPT-4.1 direct-translation performance.
    \item We show that \copit{}-generated synthetic data enables bidirectional translation, improving both Target$\rightarrow$Traditional Mongolian and Traditional Mongolian$\rightarrow$Target directions.
\end{itemize}
\section{Related Work}\label{sec:2_related_work}
\subsection{Low-Resource Language Processing}\label{subsec:2_1_lrl}
Prior work on low-resource language processing can be broadly categorized into three lines of research.
\textit{Data-centric} methods aim to expand parallel data through web mining~\citep{schwenk-etal-2021-wikimatrix, schwenk-etal-2021-ccmatrix}, back-translation~\citep{sennrich-etal-2016-improving, edunov-etal-2018-understanding}, or LLM-generated synthetic corpora~\citep{de-gibert-etal-2025-scaling, waldendorf-etal-2025-multilingual}.
\textit{Model-centric} methods leverage multilingual pretrained models~\citep{conneau-etal-2020-unsupervised, xue-etal-2021-mt5}, but their shared representation assumptions often break down for languages with divergent or multiple writing systems~\citep{liu-etal-2024-translico}. \textit{Linguistically motivated} approaches incorporate explicit knowledge such as transliteration~\citep{ma2025exploring} or morphology-aware tokenization~\citep{chaudhary-etal-2018-adapting, nzeyimana2024low}. While effective for reducing surface-level variation, these methods do not explicitly model script-level ambiguity arising from digraphy.

Our work lies at the intersection of linguistically motivated modeling and data-centric augmentation, as the proposed pivot not only resolves script-level ambiguity but also enables the creation of synthetic parallel data for low-resource translation.

\subsection{Mongolian Language Processing}\label{subsec:2_2_mongolian}
Prior work on Mongolian language processing can be broadly grouped by the scripts they address: the Cyrillic script, the Traditional script, and conversion between the two. Most studies focus on Mongolian written in the Cyrillic script, including benchmark construction \citep{zhang2024mm, yugoevaluating} and language model pre-training \citep{na2024pre}. These works typically assume input text written in a single standardized script and do not account for challenges arising from digraphia.
Research on Mongolian written in the Traditional script has largely focused on classical NLP tasks such as morphological analysis \citep{liu-etal-2020-incorporating, 11228752}, with limited connection to downstream applications or LLM-based frameworks.
Work addressing both scripts has primarily framed the problem as script conversion. Early approaches relied on rule-based methods \citep{6104727}, followed by hybrid systems combining linguistic rules with neural models \citep{na2023traditional, Na_Bao_Wang_Gao_Dulamragchaa_2024}, encoder--decoder architectures \citep{na2022deep}, and more recent neural machine translation formulations \citep{dulamragchaa2024joint}.

In this work, we treat script conversion as a structured inference step that resolves script-level ambiguity and produces a meaning-consistent intermediate representation, enabling more accurate translation from the Traditional script under digraphic low-resource conditions.

\section{\copit: Cognitive Pivot Translation}\label{sec:3_method}
\subsection{Overview}\label{subsec:3_1_overview}
Our task is to translate Mongolian text written in the Traditional script into a target language. This is challenging for current LLMs because Mongolian is a digraphic low-resource language with a strong resource imbalance across scripts. The Traditional script is severely under-resourced and orthographically ambiguous, leading to significant performance degradation compared to the Cyrillic script and making direct translation difficult.

To address this challenge, we propose \copit{}, a translation pipeline for Mongolian text written in the Traditional script. As illustrated in Figure~\ref{fig:2_pipeline}, \copit{} is cognitively motivated by how fluent Mongolian readers process the Traditional script: rather than interpreting it directly, they implicitly map it to the Cyrillic script to resolve script-induced ambiguities. 
\copit{} consists of two stages: (1) Traditional-to-Cyrillic pivoting, which includes morphological segmentation, multi-step script disambiguation, and sentence-level self-reflection; and (2) translation from the disambiguated Cyrillic representation to the target language. More details, including the prompt design, are provided in Appendix~\ref{subsec:appendix_prompts}.

\subsection{Traditional-to-Cyrillic Pivoting}\label{subsec:3_2_pivoting}
\textbf{Morphological Segmentation.} 
\begin{figure}[t]
    \centering
    \includegraphics[width=0.95\columnwidth]{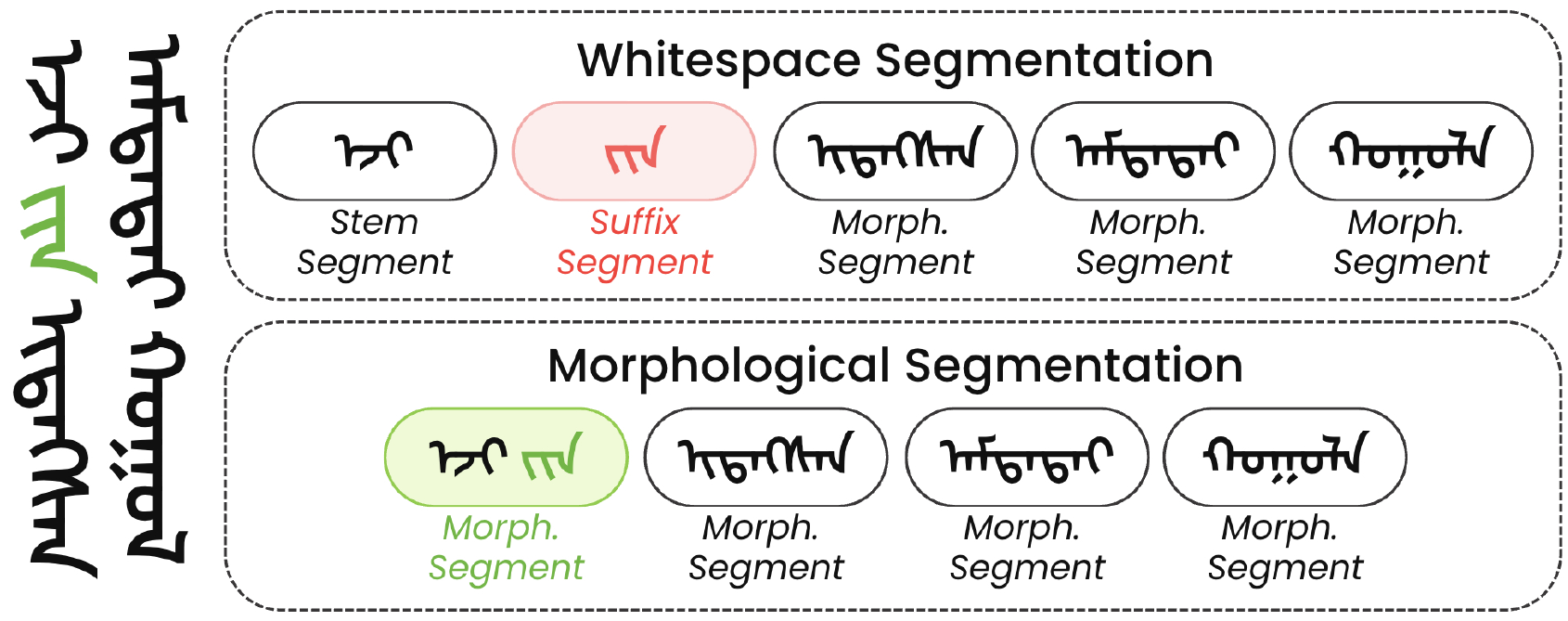}
    \caption{Comparison between whitespace-based segmentation and morphological segmentation for Mongolian sentence written in the Traditional script.}
    \label{fig:3_morphological_segmentation}
    \vspace{-1em}
\end{figure}
As illustrated in Figure~\ref{fig:3_morphological_segmentation}, the Traditional script frequently inserts spaces before suffixes rather than attaching them to stems, a convention that makes surface whitespace an unreliable indicator of true morphological boundaries~\citep{janhunen2012mong}. Morphological segmentation thus requires distinguishing true lexical boundaries from suffix attachment.

The process begins with a simple whitespace-based split to obtain candidate tokens, followed by dictionary-guided merging of stems and suffixes. A curated suffix dictionary is used to identify suffixes that should be combined with the preceding token to form a single morphological segment. These suffixes encode essential grammatical information, such as case and verbal categories. This step produces the morphologically coherent segments required for subsequent script disambiguation steps. 

\begin{figure}[t]
    \centering
    \includegraphics[width=0.95\columnwidth]{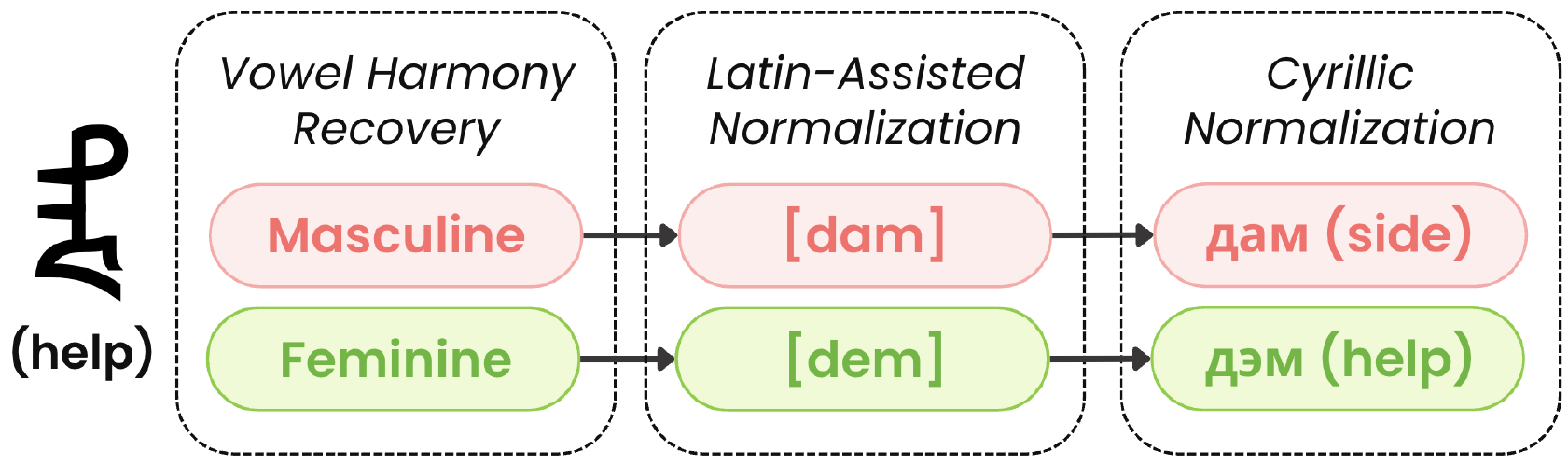}
    \caption{Multi-step script disambiguation via vowel harmony recovery and Latin-to-Cyrillic normalization.}
    \label{fig:4_script_disambiguation}
    \vspace{-1em}
\end{figure}

\noindent
\textbf{Multi-Step Script Disambiguation.}
Segmentation yields well-formed morphological units, but script-inherent ambiguity may still remain at the segment level. We address this through a structured multi-step script disambiguation process illustrated in Figure~\ref{fig:4_script_disambiguation}. The process is implemented using an LLM and progressively incorporates explicit phonological and orthographic constraints.

\textit{Vowel Harmony Recovery Step.}
Vowel harmony is a core phonological constraint in Mongolian, requiring vowels within a word to align with either a masculine or feminine harmony class~\citep{svantesson2024vowel}. In the Traditional script, medial vowel shapes often underspecify this distinction, making harmony class identification nontrivial. At this step, the harmony class of each segmented unit is inferred. The LLM determines the harmony class primarily based on the first vowel in the segment, following standard phonological rules, while accounting for neutral vowels that do not independently determine harmony. Enforcing vowel harmony eliminates interpretations that violate phonological constraints and substantially narrows the space of plausible analyses, providing a strong constraint for subsequent normalization steps.

\textit{Latin-Assisted Normalization Step.} 
This step performs explicit phonological normalization by mapping each segmented unit into an intermediate Latin-based representation that approximates its phonological realization in modern Mongolian. Rather than direct transliteration, the goal is to make phonological distinctions that are implicit in the Traditional script explicit. The mapping follows standard phonetic correspondences for vowels, diphthongs, and consonants, while leveraging the recovered vowel harmony class to disambiguate visually ambiguous medial forms. By rendering underspecified phonological structure explicit, the Latin representation substantially reduces the space of plausible interpretations and provides a clearer foundation for subsequent Cyrillic normalization.

\textit{Cyrillic Normalization Step.}
In the final script disambiguation step, each Latin-normalized segment is mapped to a canonical Cyrillic form. The Cyrillic script serves as an intermediate normalization space rather than the final translation target. Guided by the phonologically clarified Latin representation and previously inferred constraints, the LLM resolves remaining orthographic variation and selects a single, well-formed Cyrillic realization. This normalized representation provides a stable foundation for downstream translation by enabling reliable comparison among alternatives and filtering out morphologically or semantically incompatible candidates.

\noindent
\textbf{Sentence Reconstruction with Self-Reflection.}
Once local ambiguities are resolved, we reconstruct the full sentence in the Cyrillic script by assembling the previously disambiguated segments. This is necessary because translation requires a globally consistent, sentence-level representation rather than independently resolved segments.

\begin{figure}[t]
    \centering
    \includegraphics[width=0.95\columnwidth]{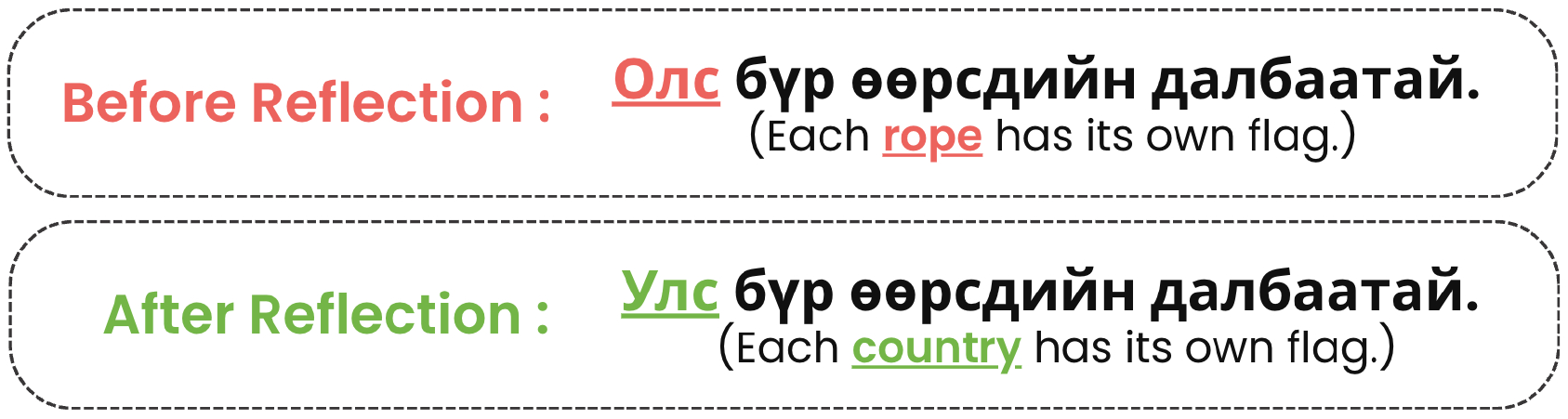}
    \caption{Sentence-level self-reflection for global semantic and grammatical consistency.}
    \label{fig:5_self_reflection}
    \vspace{-1em}
\end{figure}

We then incorporate a self-reflection component, inspired by prior work that employs self-reflection mechanisms for grammatical correction and machine translation~\citep{wang2024taste, chen2024dual}. The model reviews its earlier local decisions and revises them only when they result in semantic or grammatical inconsistencies at the sentence level, as shown in Figure~\ref{fig:5_self_reflection}. Through this process, the model selects a single globally coherent analysis for the sentence, resolving distinctions such as lemma choice, case marking, or tense interpretation that cannot be reliably determined from the surface form of the Traditional script alone. Because many ambiguities in the Traditional script cannot be resolved locally, this reflection component plays a critical role in producing a coherent and internally consistent Cyrillic representation suitable for downstream translation.

\subsection{Translation from the Disambiguated Representation}\label{subsec:3_3_translation}
Finally, the disambiguated Cyrillic representation is translated into the target languages—English, Korean, and Russian—using a prompt-tuned LLM. Crucially, the MT system no longer operates under script-induced uncertainty: the ambiguities inherent in the Traditional script have been resolved upstream, reducing error propagation into the translation output. As a result, the translation model can focus on cross-lingual meaning transfer rather than script interpretation, yielding translations that better reflect the intended semantics of the original Traditional-script sentence. Compared to direct translation, this pipeline yields more stable translation behavior without requiring additional training, by explicitly resolving script-level ambiguity and enabling intermediate consistency checks.

\subsection{Component-wise Training}\label{subsec:3_4_training}
We train each LLM-based component independently using supervised fine-tuning with parameter-efficient adaptation~\citep{hu2022lora}. Each component is trained with dedicated prompts and objectives, enabling effective specialization in its intended function within the overall pipeline.

The Vowel Harmony Recovery, Latin-Assisted Normalization, and Cyrillic Normalization components are trained using \emph{word-level} supervision from a Traditional--Cyrillic parallel lexical dataset comprising 14,125 entries. This dataset is derived from a dictionary that aligns Traditional and Cyrillic word forms with phonetic transliterations, with vowel harmony labels automatically assigned and verified by a native Mongolian speaker. Because the data consist of dictionary-level word pairs rather than sentence-level parallel corpora, they can be collected with substantially lower cost and effort, which is particularly advantageous in low-resource settings. The proposed components are explicitly designed to leverage this word-level lexical supervision and apply it compositionally to longer input sequences in the end-to-end pipeline.

The sentence-level self-reflection component is trained separately using 2,061 synthetic Cyrillic revision pairs. To construct this data, sentences written in the Traditional script are first converted into Cyrillic by a base LLM to produce imperfect or noisy hypotheses, which are then paired with their original Cyrillic counterparts. This data construction process is relatively straightforward in Mongolian due to its digraphic nature, where equivalent content is commonly available in both scripts. Training on such revision pairs enables the model to correct locally plausible but globally inconsistent analyses at the sentence level. All datasets used for component-wise training are publicly released to support reproducibility and future research.
\begin{table*}[t]
\renewcommand{\arraystretch}{0.95}
\footnotesize
\centering

\resizebox{\textwidth}{!}{%
\begin{tabular}{ll|ccc|ccc|ccc}
\toprule
\multirow{2}{*}{\textbf{Backbone}} 
& \multirow{2}{*}{\textbf{Method}}
& \multicolumn{3}{c|}{\textbf{English}}
& \multicolumn{3}{c|}{\textbf{Korean}}
& \multicolumn{3}{c}{\textbf{Russian}} \\
& 
& BLEU-3/4$\uparrow$ & chrF/chrF++$\uparrow$ & COMET$\uparrow$
& BLEU-3/4$\uparrow$ & chrF/chrF++$\uparrow$ & COMET$\uparrow$
& BLEU-3/4$\uparrow$ & chrF/chrF++$\uparrow$ & COMET$\uparrow$ \\
\midrule

\multirow{3}{*}{Qwen-3 (4B)} & \textit{Direct}
 & 0.54/0.21 & 13.34/11.73 & 0.373
 & 0.74/0.35 & 2.88/2.75 & 0.351 
 & 0.35/0.10 & 5.03/4.65 & 0.266 \\
 & \textit{CoPiT (Zero-shot)}
 & \underline{0.94}/\underline{0.46} & \underline{15.57}/\underline{13.73} & \underline{0.422}
 & \underline{0.92}/\underline{0.41} & \underline{3.50}/\underline{3.36} & \underline{0.372}
 & \underline{0.59}/\underline{0.31} & \underline{11.42}/\underline{9.37} & \underline{0.312} \\
 & \textit{CoPiT (Fine-tuned)}
 & \textbf{9.40}/\textbf{5.70} & \textbf{29.70}/\textbf{27.55} & \textbf{0.628}
 & \textbf{10.78}/\textbf{6.65} & \textbf{11.15}/\textbf{10.91} & \textbf{0.639}
 & \textbf{3.45}/\textbf{1.99} & \textbf{18.60}/\textbf{16.24} & \textbf{0.544} \\
\cmidrule(lr){1-11}

\multirow{3}{*}{Qwen-3 (30B)} & \textit{Direct}
 & 1.25/0.48 & 19.73/17.51 & 0.443
 & 1.28/0.59 & 3.94/3.85 & 0.421
 & 0.33/0.12 & 9.55/8.33 & 0.331 \\
 & \textit{CoPiT (Zero-shot)}
 & \underline{1.86}/\underline{0.92}& \underline{21.75}/\underline{19.28} & \underline{0.493}
 & \underline{2.95}/\underline{1.43} & \underline{6.08}/\underline{5.87} & \underline{0.498}
 & \underline{0.83}/\underline{0.39} & \underline{13.41}/\underline{11.54} & \underline{0.388} \\
 & \textit{CoPiT (Fine-tuned)}
 & \textbf{7.61}/\textbf{4.56} & \textbf{29.25}/\textbf{26.94} & \textbf{0.593}
 & \textbf{9.31}/\textbf{5.81} & \textbf{10.91}/\textbf{10.65} & \textbf{0.595}
 & \textbf{2.83}/\textbf{1.55} & \textbf{19.51}/\textbf{17.00} & \textbf{0.487} \\
\cmidrule(lr){1-11}

\multirow{3}{*}{Ministral-3 (3B)} & \textit{Direct}
 & \underline{1.21}/\underline{0.46} & \underline{23.19}/\underline{20.21} & \underline{0.500}
 & 0.39/0.25 & 0.65/0.68 & 0.236
 & 0.60/0.20 & 14.08/12.03 & 0.404 \\
 & \textit{CoPiT (Zero-shot)}
 & 1.09/0.22 & 22.66/19.82 & 0.497
 & \underline{0.81}/\underline{0.30} & \underline{3.61}/\underline{3.42} & \underline{0.393}
 & \underline{0.69}/\underline{0.26} & \underline{16.34}/\underline{14.15} & \underline{0.439} \\
 & \textit{CoPiT (Fine-tuned)}
 & \textbf{4.81}/\textbf{2.66} & \textbf{29.38}/\textbf{26.21} & \textbf{0.619}
 & \textbf{5.59}/\textbf{3.17} & \textbf{9.46}/\textbf{8.88} & \textbf{0.585}
 & \textbf{2.21}/\textbf{1.00} & \textbf{21.92}/\textbf{19.08} & \textbf{0.547} \\
\cmidrule(lr){1-11}

\multirow{3}{*}{Ministral-3 (14B)} & \textit{Direct}
 & 1.79/0.77 & 21.44/19.13 & 0.467
 & 1.44/0.65 & 4.53/4.30 & 0.435
 & 0.92/0.34 & 16.70/14.44 & 0.473 \\
 & \textit{CoPiT (Zero-shot)}
 & \underline{3.02}/\underline{1.40} & \underline{26.02}/\underline{23.30} & \underline{0.549}
 & \underline{3.37}/\underline{1.80} & \underline{6.93}/\underline{6.72} & \underline{0.528}
 & \underline{2.31}/\underline{1.12} & \underline{21.70}/\underline{19.05} & \underline{0.558} \\
 & \textit{CoPiT (Fine-tuned)}
 & \textbf{14.16}/\textbf{9.58} & \textbf{38.24}/\textbf{35.41} & \textbf{0.707}
 & \textbf{14.64}/\textbf{9.82} & \textbf{16.01}/\textbf{15.30} & \textbf{0.714}
 & \textbf{9.04}/\textbf{5.84} & \textbf{32.49}/\textbf{29.12} & \textbf{0.728} \\
\cmidrule(lr){1-11}

\multirow{2}{*}{GPT-4.1} & \textit{Direct}
 & 2.75/1.35 & 19.59/17.66 & 0.468
 & 2.67/1.39 & 5.30/5.04 & 0.454
 & 1.57/0.73 & 15.86/14.00 & 0.516 \\
 & \textit{CoPiT (Zero-shot)}
 & \textbf{14.07}/\textbf{9.83} & \textbf{37.29}/\textbf{24.59} & \textbf{0.663}
 & \textbf{13.37}/\textbf{9.04} & \textbf{14.59}/\textbf{14.05} & \textbf{0.667}
 & \textbf{8.71}/\textbf{5.63} & \textbf{31.84}/\textbf{28.53} & \textbf{0.694} \\
 
\bottomrule
\end{tabular}
}
\vspace*{-0.5em}
\caption{Machine translation performance across languages under reference-based evaluation.}
\label{tab:mt_results_ref}
\vspace*{-0.5em}
\end{table*}
\begin{table*}[t]
\renewcommand{\arraystretch}{0.95}
\footnotesize
\centering

\resizebox{\textwidth}{!}{%
\begin{tabular}{ll|ccc|ccc|ccc}
\toprule
\multirow{2}{*}{\textbf{Backbone}} 
& \multirow{2}{*}{\textbf{Method}}
& \multicolumn{3}{c|}{\textbf{English}}
& \multicolumn{3}{c|}{\textbf{Korean}}
& \multicolumn{3}{c}{\textbf{Russian}} \\
& 
& COMETKiwi$\uparrow$ & Adeq.$\uparrow$ & Fluen.$\uparrow$
& COMETKiwi$\uparrow$ & Adeq.$\uparrow$ & Fluen.$\uparrow$
& COMETKiwi$\uparrow$ & Adeq.$\uparrow$ & Fluen.$\uparrow$ \\
\midrule

\multirow{3}{*}{Qwen-3 (4B)} & \textit{Direct}
 & 0.078 & \underline{1.33} & 1.89
 & \underline{0.204} & \underline{1.00} & 1.33
 & \underline{0.167} & \underline{1.00} & \underline{1.00} \\
 & \textit{CoPiT (Zero-shot)}
 & \underline{0.183} & 1.00 & \underline{2.00}
 & 0.160 & \underline{1.00} & \underline{1.44} 
 & 0.124 & \underline{1.00} & \underline{1.00} \\
 & \textit{CoPiT (Fine-tuned)}
 & \textbf{0.352} & \textbf{2.44} & \textbf{3.00}
 & \textbf{0.411} & \textbf{2.56} & \textbf{3.00}
 & \textbf{0.280} & \textbf{1.22} & \textbf{1.33} \\
\cmidrule(lr){1-11}

\multirow{3}{*}{Qwen-3 (30B)} & \textit{Direct}
 & 0.204 & 1.11 & 2.33
 & 0.322 & 1.00 & 1.78
 & 0.156 & 1.00 & 1.00 \\
 & \textit{CoPiT (Zero-shot)}
 & \underline{0.250} & \underline{2.11} & \underline{3.00}
 & \underline{0.376} & \underline{2.00} & \textbf{3.00} 
 & \underline{0.211} & \underline{1.56} & \underline{2.67} \\
 & \textit{CoPiT (Fine-tuned)}
 & \textbf{0.360} & \textbf{2.33} & \textbf{3.44}
 & \textbf{0.419} & \textbf{2.67} & \underline{2.89}
 & \textbf{0.301} & \textbf{1.89} & \textbf{2.22} \\
\cmidrule(lr){1-11}

\multirow{3}{*}{Ministral-3 (3B)} & \textit{Direct}
 & \underline{0.315} & \underline{1.22} & 2.67
 & \underline{0.344} & \underline{1.00} & \underline{1.44}
 & 0.228 & \underline{1.11} & \underline{1.33} \\
 & \textit{CoPiT (Zero-shot)}
 & 0.291 & \underline{1.22} & \underline{2.89}
 & 0.272 & \underline{1.00} & 1.22
 & \underline{0.237} & 1.00 & 1.00 \\
 & \textit{CoPiT (Fine-tuned)}
 & \textbf{0.402} & \textbf{2.78} & \textbf{3.22}
 & \textbf{0.413} & \textbf{2.33} & \textbf{2.56}
 & \textbf{0.338} & \textbf{1.89} & \textbf{2.22} \\
\cmidrule(lr){1-11}

\multirow{3}{*}{Ministral-3 (14B)} & \textit{Direct}
 & \underline{0.360} & 1.56 & \underline{2.44}
 & 0.391 & 1.00 & 1.44
 & \underline{0.360} & 1.00 & 1.33 \\
 & \textit{CoPiT (Zero-shot)}
 & 0.309 & \underline{2.11} & 2.33
 & \underline{0.400} & \underline{1.89} & \underline{1.89} 
 & 0.349 & \underline{1.89} & \underline{2.11} \\
 & \textit{CoPiT (Fine-tuned)}
 & \textbf{0.461} & \textbf{4.22} & \textbf{4.44}
 & \textbf{0.490} & \textbf{3.56} & \textbf{3.33}
 & \textbf{0.449} & \textbf{3.67} & \textbf{3.89} \\
\cmidrule(lr){1-11}

\multirow{2}{*}{GPT-4.1} & \textit{Direct}
 & 0.270 & 1.44 & 2.67
 & 0.249 & 1.44 & 1.56 
 & \textbf{0.429} & 1.44 & 2.67 \\
 & \textit{CoPiT (Zero-shot)}
 & \textbf{0.470} & \textbf{4.33} & \textbf{4.56}
 & \textbf{0.472} & \textbf{3.44} & \textbf{3.33}
 & 0.399 & \textbf{4.00} & \textbf{4.44} \\

\bottomrule
\end{tabular}
}
\vspace*{-0.5em}
\caption{Machine translation performance across languages under reference-free evaluation.}
\label{tab:mt_results_noref}
\vspace*{-1em}
\end{table*}

\section{Experiments}\label{sec:4_experiments}
We conduct our experiments to address the following three research questions, which collectively examine translation quality, component-level effectiveness, and the potential of the proposed pipeline to alleviate low-resource challenges.
\begin{table*}[t]
\renewcommand{\arraystretch}{0.95}
\centering

\resizebox{\textwidth}{!}{%
\begin{tabular}{cl|ccc|ccc|ccc}
\toprule
\multirow{2}{*}{\textbf{Backbone}} 
& \multirow{2}{*}{\textbf{Method}}
& \multicolumn{3}{c|}{\textbf{English}}
& \multicolumn{3}{c|}{\textbf{Korean}}
& \multicolumn{3}{c}{\textbf{Russian}} \\
& 
& BLEU-3/4$\uparrow$ & chrF/chrF++$\uparrow$ & COMET$\uparrow$
& BLEU-3/4$\uparrow$ & chrF/chrF++$\uparrow$ & COMET$\uparrow$
& BLEU-3/4$\uparrow$ & chrF/chrF++$\uparrow$ & COMET$\uparrow$ \\
\midrule

\multirow{5.5}{*}{\makecell{Qwen-3\\(4B)}} & \textit{Direct}
 & 0.54/0.21 & 13.34/11.73 & 0.373
 & 0.74/0.35 & 2.88/2.75 & 0.351 
 & 0.35/0.10 & 5.03/4.65 & 0.266 \\
 
 & \textit{CoPiT w/o VHR}
 & \underline{8.08}/\underline{5.00} & \underline{29.57}/\textbf{29.37} & \textbf{0.633}
 & \underline{6.79}/\underline{4.17} & \underline{10.50}/\underline{10.23} & \underline{0.638} 
 & \underline{2.00}/\underline{1.12} & \underline{17.31}/\underline{15.17} & \textbf{0.546} \\
 
 & \textit{CoPiT w/o LAN}
 & 3.06/1.55 & 23.34/21.28 & 0.568
 & 3.42/1.93 & 7.25/7.09 & 0.558
 & 0.74/0.39 & 13.04/11.20 & 0.456 \\

 & \textit{CoPiT w/o SR}
 & 1.44/0.74 & 18.22/16.36 & 0.550
 & 1.60/0.88 & 5.19/4.99 & 0.540 
 & 0.44/0.20 & 10.50/8.94 & 0.415 \\
 \cmidrule(lr){2-11}

 & \textit{CoPiT}
 & \textbf{9.40}/\textbf{5.70} & \textbf{29.70}/\underline{27.55} & \underline{0.628}
 & \textbf{10.78}/\textbf{6.65} & \textbf{11.15}/\textbf{10.91} & \textbf{0.639}
 & \textbf{3.45}/\textbf{1.99} & \textbf{18.60}/\textbf{16.24} & \underline{0.544} \\
\cmidrule(lr){1-11}

\multirow{5.5}{*}{\makecell{Ministral-3\\(3B)}} & \textit{Direct}
 & 1.21/0.46 & 23.19/20.21 & 0.500
 & 0.39/0.25 & 0.65/0.68 & 0.236
 & 0.60/0.20 & 14.08/12.03 & 0.404 \\
 
 & \textit{CoPiT w/o VHR}
 & \textbf{5.02}/\textbf{2.71} & \textbf{29.92}/\textbf{26.72} & \textbf{0.625}
 & \underline{4.57}/\underline{2.66} & \underline{9.17}/\underline{8.59} & \textbf{0.594}
 & \underline{2.03}/\underline{1.07} & \underline{20.30}/\underline{17.62} & \underline{0.542} \\
 
 & \textit{CoPiT w/o LAN}
 & 3.86/1.95 & 28.64/25.47 & 0.607
 & 3.40/1.89 & 7.78/7.30 & 0.563
 & 1.60/0.84 & 18.60/16.12 & 0.507 \\
 
 & \textit{CoPiT w/o SR}
 & 2.46/1.16 & 26.41/23.21 & 0.577
 & 1.68/0.82 & 5.84/5.41 & 0.510
 & 0.58/0.22 & 15.17/13.04 & 0.442 \\
 \cmidrule(lr){2-11}

 & \textit{CoPiT}
 & \underline{4.81}/\underline{2.66} & \underline{29.38}/\underline{26.21} & \underline{0.619}
 & \textbf{5.59}/\textbf{3.17} & \textbf{9.46}/\textbf{8.88} & \underline{0.585}
 & \textbf{2.21}/\textbf{1.00} & \textbf{21.92}/\textbf{19.08} & \textbf{0.547} \\
 
\bottomrule
\end{tabular}
}

\caption{Ablation study of individual components in the proposed \copit{} across target languages under reference-based evaluation, where \textit{VHR}, \textit{LAN}, and \textit{SR} denote Vowel Harmony Recovery, Latin-Assisted Normalization, and Self-Reflection, respectively.}
\label{tab:ablation_results}
\end{table*}
\input{}

\noindent\textbf{RQ1.} 
Does our pipeline improve end-to-end machine translation performance over baselines?

\noindent\textbf{RQ2.} 
How does each component affect overall translation quality?

\noindent\textbf{RQ3.} 
Can the pipeline contribute to mitigating low-resource data challenges?

\subsection{Experimental Setups}
\textbf{Datasets.}
Due to the severe data scarcity of Mongolian, it is impractical to collect parallel corpora that are well aligned across languages or to obtain fully human written reference translations. We therefore construct two datasets for reference-based and reference-free evaluation, respectively.

The first dataset contains 1,031 sentence-level web-crawled Traditional–Cyrillic parallel pairs with human-verified reference translations. English references are translated from the Cyrillic side and subsequently validated by bilingual speakers. Korean and Russian references are then generated from the verified English sentences and independently reviewed by fluent speakers of each target language. This dataset enables reliable reference-based evaluation across multiple target languages.

The second dataset comprises 380 news articles, each containing about five sentences on average, in the Traditional and Cyrillic scripts collected from the official Mongolian president's website.\footnote{\url{https://president.mn/en/}} As no reference translations are available, this dataset is used exclusively for reference-free evaluation.

\noindent
\textbf{Backbone Models.}
We adopt Qwen-3 (4B, 30B) \citep{yang2025qwen3} and Ministral-3 (3B, 14B) \citep{mistral3-release} as state-of-the-art multilingual open-source backbones, and GPT-4.1 \citep{achiam2023gpt, openai-gpt4-1} as representative commercial LLMs, enabling a comprehensive evaluation across model scale, and real-world deployment settings.

\noindent
\textbf{Evaluation Metrics.}
We employ different evaluation metrics depending on the availability of reference translations. For datasets with references, we use BLEU-3/4~\citep{papineni2002bleu} and chrF/chrF++~\citep{popovic2015chrf, popovic2017chrf++} to measure lexical overlap, along with COMET~\citep{rei2022comet} to assess semantic adequacy.

For datasets without reference translations, we adopt COMETKiwi~\citep{rei2023scaling}, a reference-free quality estimation metric that scores translation quality from source–hypothesis pairs. We additionally conduct human evaluation with six fluent bilingual speakers of Mongolian and each target language. Translations are evaluated along two criteria: \textbf{Adequacy (Adeq.)}: Faithful meaning preservation with respect to the source text. \textbf{Fluency (Fluen.)}: Grammaticality and overall naturalness of the translation in the target language.

\noindent
\textbf{Implementation Details.}
Open-source backbone models are fine-tuned with LoRA, as described in Section~\ref{subsec:3_4_training}. GPT-4.1 is accessed via its APIs without additional fine-tuning. Additional implementation details, including hyperparameters, train/validation/test dataset split information, and inference settings, are provided in Appendix ~\ref{subsec:appendix_trainingdetails}.

\subsection{Experimental Results}\label{subsec:4_2_results}
\noindent
\subsubsection{Machine Translation Evaluation}
Table~\ref{tab:mt_results_ref} presents reference-based evaluation results across English, Korean, and Russian. Across all backbones, fine-tuned \copit{} outperforms direct translation. Even without training, the \copit{} pipeline yields measurable gains, particularly for Korean and Russian, indicating the effectiveness of script-level pivoting for typologically distant languages. For example, on Ministral-3 (3B), zero-shot \copit{} improves COMET by +0.157 for Korean and +0.035 for Russian, while yielding no improvement for English. With component fine-tuning (check, gains become substantial across all languages; on Ministral-3 (14B), COMET increases by +0.240 (English), +0.279 (Korean), and +0.255 (Russian). Similar trends are observed for BLEU and chrF/chrF++, where fine-tuned \copit{} consistently achieves higher scores across languages. Overall, larger backbones attain higher absolute performance, and fine-tuned \copit{} on open-source models matches or exceeds GPT-4.1 under direct translation.

Table~\ref{tab:mt_results_noref} reports reference-free evaluation results using COMETKiwi and human adequacy and fluency ratings. Consistent with reference-based findings, fine-tuned \copit{} improves translation quality across most backbones over direct method. On Ministral-3 (14B), fine-tuned \copit{} improves COMETKiwi by +0.101 (English), +0.099 (Korean), and +0.089 (Russian), accompanied by gains in human adequacy and fluency across most conditions. When applied to GPT-4.1, zero-shot \copit{} also yields improvements, increasing COMETKiwi by +0.200 (English) and +0.223 (Korean), highlighting the effectiveness of pivot-based inference in reference-scarce settings.
Across both reference-based and reference-free evaluations, \copit{} improves translation quality across model backbones and target languages. Applying the \copit{} pipeline alone already provides gains in most cases, while fine-tuning the pipeline components yields the best overall performance. These results demonstrate that script-level pivoting is an effective and practical approach for improving low-resource digraphic machine translation and for substantially reducing the performance gap between open-source and proprietary models.
Detailed annotation protocols with inter-annotator agreement analyses and few-shot prompting baselines are provided in Appendix~\ref{subsec:appendix_human_eval} and ~\ref{subsec:appendix_fewshot}, respectively.

\begin{table*}[t]
\renewcommand{\arraystretch}{0.95}
\footnotesize
\centering

\resizebox{\textwidth}{!}{%
\begin{tabular}{ll|cc|cc|cc}
\toprule
\multirow{2}{*}{\textbf{Backbone}} 
& \multirow{2}{*}{\textbf{Method}}
& \multicolumn{2}{c|}{\textbf{English→TM}}
& \multicolumn{2}{c|}{\textbf{Korean→TM}}
& \multicolumn{2}{c}{\textbf{Russian→TM}} \\
& 
& BLEU-3/4$\uparrow$ & chrF/chrF++$\uparrow$
& BLEU-3/4$\uparrow$ & chrF/chrF++$\uparrow$
& BLEU-3/4$\uparrow$ & chrF/chrF++$\uparrow$ \\
\midrule

\multirow{2}{*}{Qwen-3 (4B)}
& \textit{Base}
& 0.07/0.03 & 0.10/0.12
& 0.10/0.04 & 0.12/0.14
& 0.07/0.02 & 0.10/0.12 \\
& \textit{Fine-tuned}
& \textbf{5.04/3.22} & \textbf{26.03/22.61}
& \textbf{6.68/4.26} & \textbf{29.94/26.21}
& \textbf{4.43/2.68} & \textbf{25.08/21.78} \\
\cmidrule(lr){1-8}

\multirow{2}{*}{Ministral-3 (3B)}
& \textit{Base}
& 0.05/0.02 & 1.02/0.80
& 0.04/0.01 & 3.18/2.42
& 0.04/0.02 & 0.19/0.18 \\
& \textit{Fine-tuned}
& \textbf{5.31/3.28} & \textbf{23.81/20.64}
& \textbf{6.99/4.39} & \textbf{28.38/24.88}
& \textbf{5.42/3.36} & \textbf{24.34/21.11} \\

\bottomrule
\end{tabular}
}
\vspace*{-0.5em}
\caption{Reverse-direction translation performance across languages using \copit{}-generated synthetic parallel data.}
\label{tab:reverse_translation}
\vspace*{-1em}
\end{table*}










\noindent

\subsubsection{Ablation Study}\label{subsec:4_2_2_ablation1}
Table~\ref{tab:ablation_results} presents an ablation study of of individual components in the proposed \copit{} under reference-based evaluation. Across both Qwen-3 (4B) and Ministral-3 (3B), the full pipeline achieves the strongest overall performance, attaining the best or second-best results in most settings. This indicates that backbones benefit from the complete pipeline rather than from any single component in isolation. At the same time, performance does not consistently decrease when individual components are removed, suggesting non-trivial interactions among components and backbone-dependent sensitivities.

The impact of Vowel Harmony Recovery, in particular, varies across backbones and languages. For Qwen-3 (4B), its removal reduces BLEU scores, while chrF++ and COMET remain comparable or slightly improved. In contrast, for Ministral-3 (3B), removing Vowel Harmony Recovery improves English performance across all metrics and yields competitive or higher COMET scores for Korean, suggesting partial redundancy with the backbone's internal phonological representations. In comparison, Self-Reflection emerges as the most consistently important component: its removal leads to substantial performance degradation across both backbones, all target languages, and all evaluation metrics. Latin-Assisted Normalization provides complementary but moderate gains, indicating a supporting role in orthographic stabilization rather than a dominant contribution. Overall, these findings show that the full \copit{} pipeline improves performance across different model backbones and all target languages. The single- and pairwise- component ablation analyses for Traditional-to-Cyrillic pivoting stage is provided in Appendix~\ref{subsec:appendix_ablation}.

\noindent
\subsubsection{Low-Resource Mitigation}
\textbf{Synthetic Data Generation.} Using the proposed \copit{} pipeline with Ministral-3 (14B), we construct a new parallel corpus of \textit{8,034 sentence pairs} aligning Mongolian written in the Traditional script with English, Korean, and Russian. We clarify that this dataset is not purely synthetic; while the target-language translations are generated by our pipeline, the Traditional Mongolian source texts are curated from real-world corpora. This resource addresses a critical data gap for digraphic low-resource languages. We validate its usefulness by reversing the dataset for reverse-direction translation and training the translation module separately. All datasets are publicly released to facilitate future research. A failure mode analysis of the \copit{}-generated synthetic corpus is provided in Appendix~\ref{subsec:appendix_errors}.

\begin{figure}[t]
    \centering
    \begin{subfigure}[t]{0.48\columnwidth}
        \centering
        \includegraphics[width=\linewidth]{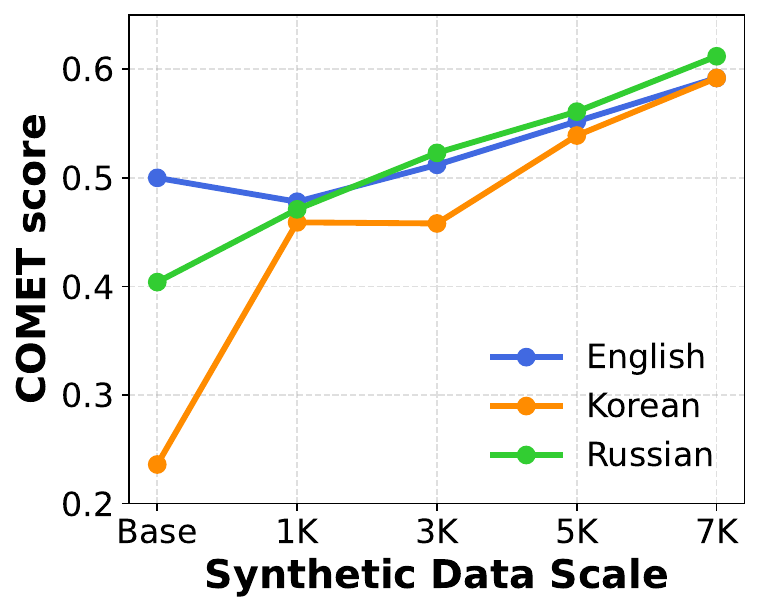}
        \caption{COMET score}
        \label{fig:6a_comet}
    \end{subfigure}
    \hfill
    \begin{subfigure}[t]{0.48\columnwidth}
        \centering
        \includegraphics[width=\linewidth]{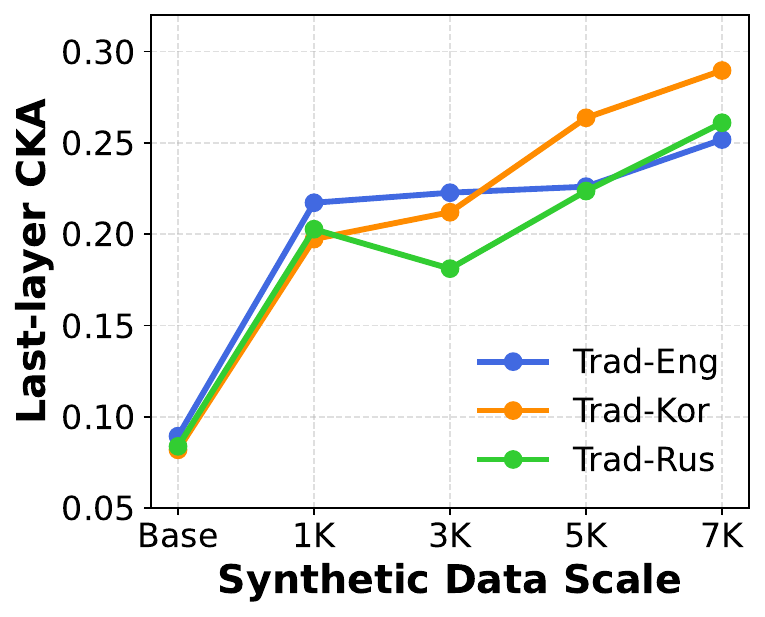}
        \caption{CKA similarity}
        \label{fig:6b_cka}
    \end{subfigure}
    \caption{Evaluation of Ministral-3 (3B) trained with varying amounts of \copit-generated synthetic data.}
    \label{fig:6_synthetic_data}
    \vspace{-1em}
\label{fig:6_synthetic}
\end{figure}

\noindent
\textbf{Data Scaling Analysis.} Figure~\ref{fig:6a_comet} presents translation performance as a function of synthetic data scale. Specifically, models were fine-tuned with Ministral-3 (3B) across increasing data scales (1K--7K) and evaluated on a fixed reference test set. Across all target languages, performance generally improves as more synthetic data is added.
From the smallest to the largest data scale, COMET scores increase by approximately $+0.09$ for English, $+0.36$ for Korean, and $+0.21$ for Russian.
At smaller scales, English shows a slight performance decrease from Base to 1K, with a reduction of approximately $-0.03$, while Korean exhibits nearly negligible changes between 1K and 3K.
Such behavior is characteristic of unstable learning dynamics in extremely low-resource regimes \citep{sennrich2019revisiting}.
Beyond this range, performance improvements remain stable across languages, despite typological differences.

To relate performance trends to internal representations, we analyze last-layer representations using centered kernel alignment (CKA) \citep{kornblith2019similarity}, which quantifies how representations induced by Traditional-script Mongolian align across different target languages as training data scale increases, as shown in Figure~\ref{fig:6b_cka}.
Overall, representations evolve smoothly as training data increases, with CKA similarity rising by roughly $+0.16$--$0.20$ across all target languages from the lowest to the highest scale.
A mild decrease in CKA similarity is observed for Russian between 1K and 3K, with an approximate reduction of $-0.02$, aligning with the low-data regime where performance gains remain limited for other target languages.
Together with the performance trends in Figure~\ref{fig:6a_comet}, this behavior is consistent with a transitional adaptation phase under extremely limited data.
In this regime, representation and performance do not yet evolve synchronously across languages.
As more data is introduced, both trends stabilize, demonstrating that CoPiT-generated data provides reliable supervision for low-resource machine translation.

\begin{figure}[t]
\centering
\includegraphics[width=\columnwidth]{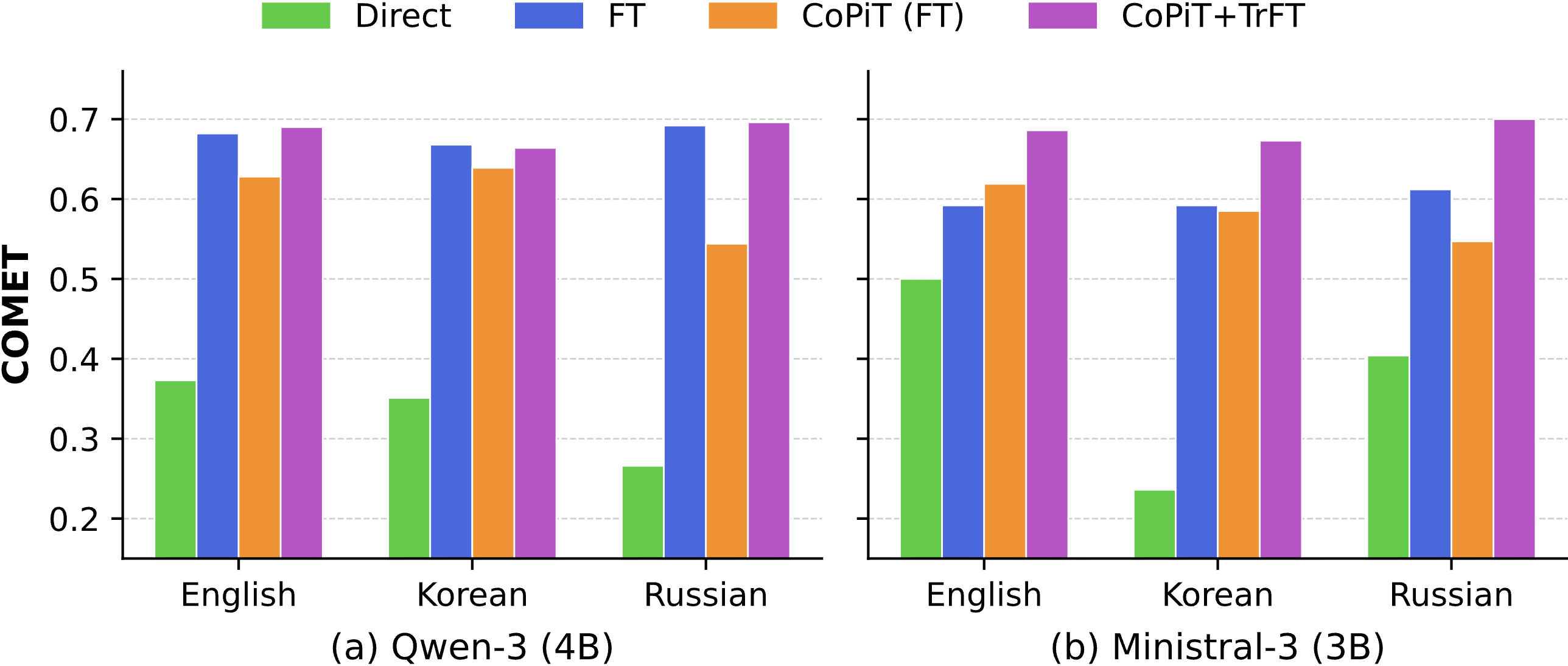}
\caption{COMET scores under pipeline configurations
with and without translation module fine-tuning.}
\label{fig:trft_half}
\end{figure}

\noindent
\textbf{Improving Forward Translation.} 
To test whether CoPiT-generated synthetic data improves forward translation, we fine-tune the translation module on the Cyrillic--Target portion of the corpus and combine it with the pivot mechanism (\textit{CoPiT+TrFT}). 
We compare with \textit{Direct} (direct translation), \textit{FT} (direct translation fine-tuned on the synthetic data), and \textit{CoPiT (FT)} (component fine-tuned \copit{} pipeline without translation module fine-tuning).
Since FT and CoPiT+TrFT are trained on identical synthetic data, performance differences generally reflect the structural contribution of the \copit{} pipeline rather than data exposure.

As shown in Figure~\ref{fig:trft_half}, CoPiT+TrFT tends to achieve strong COMET scores across most languages and backbones, suggesting that \copit{}-generated synthetic data can provide effective supervision for translation; full results across BLEU-4 and chrF++ are provided in Appendix~\ref{subsec:appendix_trft_full}.
In most settings, CoPiT+TrFT outperforms FT despite using the same training data, offering evidence that the pivot mechanism contributes independently of the data; CoPiT (FT) also performs comparably to FT in several settings, suggesting that the structural benefits of pivoting are substantial even without translation module fine-tuning with parallel data.
We note that this comparison represents an upper-bound analysis, as CoPiT is by design trained using only Traditional Mongolian--Cyrillic data without access to target-language references.

\noindent
\textbf{Reverse-Direction Translation.} To further demonstrate the practical utility of \copit{}, we examine whether the synthetic parallel data it generates can enable \textit{Target → Traditional Mongolian} translation, a direction that is otherwise infeasible or limited due to the lack of parallel resources. We construct reverse-direction training data by flipping each pair in the \copit{}-generated synthetic corpus and fine-tune the models on this reversed data.

The fine-tuned models yield generally consistent improvements over base models across languages and backbones. As shown in Table~\ref{tab:reverse_translation}, BLEU and chrF scores increase steadily for ENG → TM, KOR → TM, and RUS → TM, suggesting that the \copit{}-generated synthetic data can provide effective supervision for generating Traditional Mongolian for both backbones.

We note that evaluation for Traditional Mongolian remains challenging due to the absence of specialized metrics and the limited representation of the script in existing multilingual models. As a result, semantic metrics such as COMET are unreliable for this language. Following prior work on low-resource and morphologically rich languages, we therefore report BLEU and chrF/chrF++ scores as reference-based evaluation metrics. While our forward-direction results are verified via human evaluation, we restrict our reverse-direction analysis to the aforementioned metrics, leaving comprehensive human validation of generated Traditional Mongolian script to future work.
\section{Contribution}\label{sec:5_contribution}

In this paper, we introduce \copit{}, a cognitively motivated translation pipeline for digraphic low-resource languages, focusing on Mongolian in the Traditional script. Rather than treating Traditional-to-Cyrillic conversion as standalone transcription, \copit{} formulates it as structured disambiguation, using Cyrillic as a meaning-consistent intermediate representation for downstream translation. This design mirrors native reading practices and decouples script-level ambiguity from semantic transfer. Our pipeline integrates linguistically grounded components, vowel harmony recovery, intermediate normalization, and sentence-level reflection, each targeting a distinct source of underspecification in the Traditional script. Extensive experiments across multiple backbones and target languages show consistent improvements, including in zero-shot settings. Beyond forward translation, \copit{}-generated synthetic data enables reverse-direction translation and further improves forward translation quality via translation module fine-tuning, demonstrating the practical utility of our approach for digraphic low-resource machine translation.

\section*{Limitations}
The effectiveness of the proposed \copit{} pipeline depends in part on the quality of the downstream Cyrillic-to-target translation stage. This reliance is partly mitigated by the relatively higher availability of Traditional–Cyrillic parallel data and by the steady increase in resources, which support continued improvement of the pivoting and synthetic data generation process. As shown in Section~\ref{subsec:4_2_results}, the resulting \copit{}-generated synthetic data can further improve downstream translation quality through fine-tuning. In addition, the multi-stage design introduces higher inference latency compared to single-stage models, as multiple conversion and translation components are sequentially applied. Detailed latency analysis is provided in Appendix~\ref{subsec:appendix_inference}. Although this limits applicability to strict real-time settings, the pipeline is more suitable for offline document translation and batch-oriented low-resource MT scenarios. Finally, the core \copit{} inference pipeline is designed for outbound translation from Traditional Mongolian into non-Mongolian target languages. While our experiments show that \copit{}-generated synthetic data can support preliminary Target→Traditional Mongolian translation across multiple language pairs and backbones, a fully integrated bidirectional pipeline with dedicated evaluation metrics and human validation for Traditional Mongolian remains future work.

\section*{Acknowledgements}
This work was supported by the Institute of Information \& Communications Technology Planning \& Evaluation (IITP) grant funded by the Korean government (MSIT) (No. IITP-2026RS-2020-II201819, the ICT Creative Consilience Program, 30\%; No. RS-2026-25585074, Development of Agentic AI Technology for Enterprise-wide Business Innovation Based on Goal-Oriented Long-term Context Reasoning, 50\%; No. RS-2026-25507543, Development of AI Co-Scientist based on Scientific Causal World Model, 10\%). This work was also partly supported by the National Research Foundation of Korea (NRF) grant funded by the Korean government (MSIT) (No. RS-2025-24533089, 10\%). 

\bibliography{custom}

\clearpage
\appendix
\section{Appendix}\label{sec:appendix}
\subsection{Content Filtering in Traditional Script}\label{subsec:appendix_gemini}
\begin{figure}[t]
    \centering
    \includegraphics[width=0.95\columnwidth]{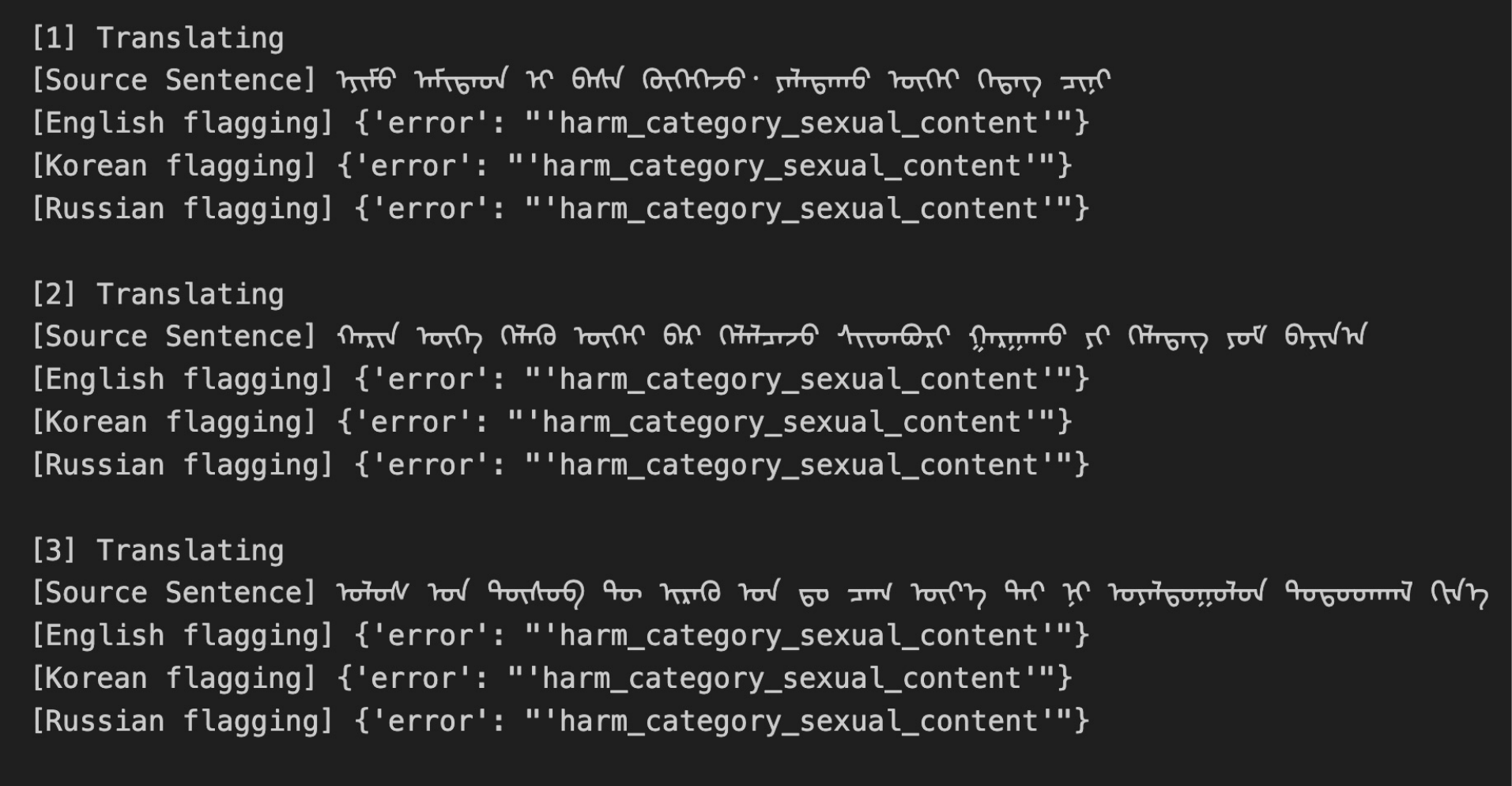}
    \caption{Example of a translation request flagged by the safety system.}
    \label{fig:7_gemini_api}
\end{figure}

Figure~\ref{fig:7_gemini_api} illustrates cases in which the Gemini-3-Pro-Preview API classifies non-offensive inputs written in the Traditional Mongolian script as harmful, resulting in blocked translation outputs. This behavior highlights limitations in script-level content filtering, as discussed in Section~\ref{sec:1_introduction}.

\subsection{Implementation and Training Details}\label{subsec:appendix_trainingdetails}

All experiments are conducted on two NVIDIA RTX PRO 6000 GPUs using CUDA 13.0. For open-source backbone models, we apply LoRA-based parameter-efficient fine-tuning with a batch size of 4, gradient accumulation of 2--4 steps, a learning rate of $1\times10^{-4}$ with a cosine scheduler, and train the models for 2--3 epochs. Minor variations within these ranges are introduced only to accommodate differences in model scale, while the overall optimization strategy and evaluation protocol remain identical across models. For inference, we consistently use greedy decoding for all models to ensure deterministic and reproducible outputs.

Table~\ref{tab:data_splits} summarizes the train/validation/test splits, including their data sources, purposes, and relevant sections. Each component is trained independently on its designated training set, with model selection based on the corresponding validation set.

\subsection{Human Evaluation}\label{subsec:appendix_human_eval}

We conduct a human evaluation on three articles, comprising a total of 12 sentences. The evaluation results are reported in Table~\ref{tab:mt_results_noref}. For each target language, six annotators are assigned. All annotators are native Mongolian speakers who are proficient in both Traditional Mongolian and Cyrillic Mongolian scripts, as well as in the corresponding target languages.

Annotators were provided with written instructions describing the evaluation task and rating criteria. They were informed that the study evaluates machine translation quality for research purposes only and that no personally identifiable information would be collected. Participation was voluntary.

Annotators evaluate translations using a 5-point Likert scale along two dimensions: Adequacy and Fluency. The detailed criteria for each score are defined as follows.
\noindent
\textit{Adequacy (Adeq.)} measures how well the translation preserves the meaning of the source text.
\begin{itemize}
\item \textbf{5 (Excellent)}: The source meaning is fully preserved with no omissions or distortions.
\item \textbf{4 (Good)}: The main meaning is preserved with only minor, non-critical errors.
\item \textbf{3 (Fair)}: The translation partially preserves the meaning but contains noticeable errors or omissions.
\item \textbf{2 (Poor)}: Only a limited portion of the source meaning is preserved.
\item \textbf{1 (Very Poor)}: The translation fails to convey the source meaning or is largely unrelated.
\end{itemize}
\noindent
\textit{Fluency (Fluen.)} assesses the grammatical correctness and naturalness of the translated text.
\begin{itemize}
\item \textbf{5 (Excellent)}: The translation is fully fluent and natural, resembling native-written text.
\item \textbf{4 (Good)}: The translation is mostly fluent with minor grammatical or stylistic issues.
\item \textbf{3 (Fair)}: The translation is generally understandable but contains noticeable grammatical or phrasing errors.
\item \textbf{2 (Poor)}: The translation is disfluent with frequent errors that hinder readability.
\item \textbf{1 (Very Poor)}: The translation is incomprehensible or severely ungrammatical.
\end{itemize}

\begin{table}[t]
\centering

\resizebox{\columnwidth}{!}{
\begin{tabular}{llclc}
\toprule
\textbf{Phase} & \textbf{Data Source} & \textbf{Train / Val / Test} & \textbf{Purpose} & \textbf{Sec.} \\
\midrule
Training & Word-level lexical pairs & 14{,}125 / 3{,}532 / -- & VHR, LAN training & 3.4 \\
Training & Sentence-level revision pairs & 2{,}061 / 343 / -- & SR training & 3.4 \\
Evaluation & Sentence-level parallel pairs & -- / -- / 1{,}031 & Reference-based evaluation & 4.1 \\
Evaluation & News articles & -- / -- / 380 & Reference-free evaluation & 4.1 \\
\bottomrule
\end{tabular}
}

\caption{Data splits used for training and evaluation.}
\label{tab:data_splits}
\end{table}

\begin{table}[t]
\renewcommand{\arraystretch}{0.95}
\footnotesize
\centering

\resizebox{\columnwidth}{!}{
\begin{tabular}{cccc}
\toprule
\multirow{2.5}{*}{\textbf{Language}} 
& \multirow{2.5}{*}{\textbf{\#Annotators}} 
& \multirow{2.5}{*}{\textbf{\#Systems}} 
& \textbf{Krippendorff's $\alpha$} \\
\cmidrule(lr){4-4}
& & & \textbf{Adeq.} \,/\, \textbf{Fluen.} \\
\midrule
English  & 6 & 14 & 0.630 / 0.263 \\
Korean  & 6 & 14 & 0.639 / 0.274 \\
Russian & 6 & 14 & 0.618 / 0.635 \\
\bottomrule
\end{tabular}
}

\caption{Inter-annotator agreement (Krippendorff’s $\alpha$)}
\label{tab:human_alpha}
\end{table}

\begin{table}[t]
\centering

\resizebox{\columnwidth}{!}{
\begin{tabular}{lcccc}
\toprule
\textbf{Language} & \textbf{Combined $\alpha$} & \textbf{B1 $\alpha$ / B2 $\alpha$} & \textbf{Pairwise $|\Delta|\leq1$} \\
\midrule
English & 0.263 & 0.378 / 0.200 & \textbf{73\%} \\
Korean & 0.274 & 0.578 / 0.135 & \textbf{78\%} \\
Russian & 0.635 & 0.553 / 0.738 & \textbf{81\%} \\
\bottomrule
\end{tabular}
}
\caption{Fluency agreement statistics.}
\label{tab:fluency_agreement}
\end{table}
Inter-annotator agreement is measured using Krippendorff’s $\alpha$~\citep{krippendorff2019}, computed separately for adequacy and fluency (see Table~\ref{tab:human_alpha}). Adequacy shows moderate agreement across languages ($\alpha=0.618$--$0.639$), indicating reasonable consistency in meaning preservation judgments. Fluency exhibits greater variability across languages (English: $\alpha=0.263$; Korean: $\alpha=0.274$; Russian: $\alpha=0.635$), reflecting the subjective nature of fluency assessment, as perceptions of naturalness vary across annotators.
To further examine this fluency pattern, we conducted a per-batch analysis. Agreement shown in Table \ref{tab:fluency_agreement} varies across groups: some batches show moderate consistency (e.g., Korean B1; Russian B1 and B2), while others are lower (e.g., English B2; Korean B2). This variability suggests differences in scoring tendencies, where certain annotators systematically use lower or higher portions of the rating scale.

Importantly, absolute difference analysis indicates that disagreements are predominantly small: 73\% (English), 78\% (Korean), and 81\% (Russian) of pairwise ratings are exact matches or differ by only one level, while extreme disagreements are rare. This pattern suggests boundary-level variation in fluency judgments.

Overall, although fluency assessment involves inherent subjectivity and scale variation, the structured disagreement pattern and limited extreme divergence support the use of the annotations for comparative system analysis.

\subsection{Few-shot Prompting Evaluation}\label{subsec:appendix_fewshot}
Table~\ref{tab:mt_results_fs} presents machine translation performance under few-shot prompting. Few-shot baselines are additionally evaluated for Ministral-3 (3B) and Qwen-3 (4B) under both Direct and \copit{} settings; results for other conditions are carried over from Table~\ref{tab:mt_results_ref} for reference. Both \copit{} and Direct improve over their respective zero-shot counterparts with few-shot prompting; however, fine-tuned \copit{} still achieves the best performance across languages and backbones, indicating that supervised adaptation drives the primary gains rather than prompt-level context alone.
\begin{table*}[t]
\renewcommand{\arraystretch}{0.95}
\footnotesize

\centering

\resizebox{\textwidth}{!}{%
\begin{tabular}{ll|ccc|ccc|ccc}
\toprule
\multirow{2}{*}{\textbf{Backbone}} 
& \multirow{2}{*}{\textbf{Method}}
& \multicolumn{3}{c|}{\textbf{English}}
& \multicolumn{3}{c|}{\textbf{Korean}}
& \multicolumn{3}{c}{\textbf{Russian}} \\
& 
& BLEU-3/4$\uparrow$ & chrF/chrF++$\uparrow$ & COMET$\uparrow$
& BLEU-3/4$\uparrow$ & chrF/chrF++$\uparrow$ & COMET$\uparrow$
& BLEU-3/4$\uparrow$ & chrF/chrF++$\uparrow$ & COMET$\uparrow$ \\
\midrule

\multirow{5}{*}{Qwen-3 (4B)} & \textit{Direct}
 & 0.54/0.21 & 13.34/11.73 & 0.373
 & 0.74/0.35 & 2.88/2.75 & 0.351 
 & 0.35/0.10 & 5.03/4.65 & 0.266 \\
  & \textit{Direct (Few-shot)}
 & \underline{1.78}/\underline{0.78} & 19.45/17.10 & \underline{0.491}
 & \underline{3.25}/\underline{1.59} & \underline{4.79}/\underline{4.66} & \underline{0.505}
 & 0.72/0.34 & 13.95/11.65 & \underline{0.486} \\
    & \textit{CoPiT (Zero-shot)}
 & 0.94/0.46 & 15.57/13.73 & 0.422
 & 0.92/0.41 & 3.50/3.36 & 0.372
 & 0.59/0.31 & 11.42/9.37 & 0.312 \\
  & \textit{CoPiT (Few-shot)}
 & 1.70/\underline{0.78} & \underline{20.01}/\underline{17.58} & 0.463
 & 1.89/0.84 & 4.78/4.60 & 0.434
 & \underline{0.91}/\underline{0.39} & \underline{15.42}/\underline{13.07} & 0.397 \\
 & \textit{CoPiT (Fine-tuned)}
 & \textbf{9.40}/\textbf{5.70} & \textbf{29.70}/\textbf{27.55} & \textbf{0.628}
 & \textbf{10.78}/\textbf{6.65} & \textbf{11.15}/\textbf{10.91} & \textbf{0.639}
 & \textbf{3.45}/\textbf{1.99} & \textbf{18.60}/\textbf{16.24} & \textbf{0.544} \\
\cmidrule(lr){1-11}

\multirow{5}{*}{Ministral-3 (3B)} & \textit{Direct}
 & \underline{1.21}/\underline{0.46} & 23.19/20.21 & 0.500
 & 0.39/0.25 & 0.65/0.68 & 0.236
 & 0.60/0.20 & 14.08/12.03 & 0.404 \\
   & \textit{Direct (Few-shot)}
 & 1.15/0.43 & 16.72/14.39 & 0.456
 & \underline{2.56}/\underline{1.05} & 4.65/4.32 & 0.474
 & \underline{0.96}/\underline{0.39} & 17.77/14.94 & 0.511 \\
  & \textit{CoPiT (Zero-shot)}
 & 1.09/0.22 & 22.66/19.82 & 0.497
 & 0.81/0.30 & 3.61/3.42 & 0.393
 & 0.69/0.26 & 16.34/14.15 & 0.439 \\
   & \textit{CoPiT (Few-shot)}
 & 1.12/0.38 & \underline{23.54}/\underline{20.37}& \underline{0.514}
 & 2.33/0.99 & \underline{5.68}/\underline{5.38} & \underline{0.491}
 & 0.89/0.17 & \underline{21.45}/\underline{18.34} & \underline{0.51}7 \\
 & \textit{CoPiT (Fine-tuned)}
 & \textbf{4.81}/\textbf{2.66} & \textbf{29.38}/\textbf{26.21} & \textbf{0.619}
 & \textbf{5.59}/\textbf{3.17} & \textbf{9.46}/\textbf{8.88} & \textbf{0.585}
 & \textbf{2.21}/\textbf{1.00} & \textbf{21.92}/\textbf{19.08} & \textbf{0.547} \\
\bottomrule

 
\end{tabular}
}

\caption{Machine translation performance under few-shot prompting, evaluated under reference-based metrics.}
\label{tab:mt_results_fs}
\end{table*}

\noindent
\subsection{Inference Latency Analysis}\label{subsec:appendix_inference}
Although \copit{} improves translation quality, it incurs additional inference latency due to its multi-stage pipeline. Table~\ref{tab:latency_comparison} shows that \copit{} requires approximately 1.9 seconds more per sample than the direct baseline, primarily due to the Vowel Harmony Recovery, Latin-Assisted Normalization, and Self-Reflection components. In contrast, the direct approach often produces substantially lower-quality translations, whereas \copit{} consistently generates higher-quality outputs across model backbones and target languages, as reflected in Table~\ref{tab:mt_results_ref}. This additional latency may be acceptable in offline or batch-oriented translation settings where translation quality is prioritized.
\begin{table}[t]
\centering
\resizebox{\columnwidth}{!}{
\begin{tabular}{lcccc}
\toprule
Method & \textit{VHR, LAN} & \textit{Self-Reflection} & \textit{Translation} & Total \\
\midrule
\textit{Direct} & -- & -- & 3.68 s & \textbf{3.68 s} \\
\textit{CoPiT} & 4.29 s & 0.67 s & 0.64 s & \textbf{5.60 s} \\
\bottomrule
\end{tabular}
}
\caption{Inference latency comparison between \textit{Direct} and \copit{}. \textit{VHR, LAN} stand for Vowel Harmony Recovery, Latin-Assisted Normalization, respectively.}
\label{tab:latency_comparison}
\end{table}

\subsection{Component Ablation Analysis}\label{subsec:appendix_ablation}
We present two complementary ablation analyses targeting different stages of the \copit{} pipeline. The first evaluates the contribution of individual components to end-to-end translation quality, covering the full pipeline including the translation stage in Section~\ref{subsec:4_2_2_ablation1}. The second focuses exclusively on the Traditional-to-Cyrillic pivoting stage, examining how component interactions affect script conversion quality prior to translation.

\begin{table}[t]
\renewcommand{\arraystretch}{1.05}
\footnotesize
\centering

\resizebox{\columnwidth}{!}{
\begin{tabular}{cl|lcc}
\toprule
\textbf{Model} 
& \textbf{Method}
& \textbf{chrF / chrF++$\uparrow$}
& \textbf{CER$\downarrow$} \\
\midrule

\multirow{3}{*}{\makecell{Qwen-3 (4B)\\+ \copit}}
& \textit{w/o VHR}  & 81.01/77.20 & 0.10 \\
& \textit{w/o LAT} & 61.85/55.71 & 0.23 \\
& \textit{w/o SR}  & 51.20/43.71 & 0.27 \\
\midrule

\multirow{3}{*}{\makecell{Ministral-3 (3B)\\+ \copit}}
& \textit{w/o VHR}  & 78.98/74.66 & 0.11 \\
& \textit{w/o LAT} & 72.50/ 67.58 & 0.16 \\
& \textit{w/o SR}  &  51.92/ 44.31 & 0.26 \\
\bottomrule
\end{tabular}
}

\caption{Ablation study of single components for Traditional-to-Cyrillic script pivoting, where \textit{VHR}, \textit{LAN}, and \textit{SR} denote Vowel Harmony Recovery, Latin-Assisted Normalization, and Self-Reflection, respectively.}

\label{tab:t2c_single_ablation}
\end{table}

\begin{table}[t]
\renewcommand{\arraystretch}{1.05}
\footnotesize
\centering

\resizebox{\columnwidth}{!}{
\begin{tabular}{cl|lcc}
\toprule
\textbf{Model} 
& \textbf{Method}
& \textbf{chrF / chrF++$\uparrow$}
& \textbf{CER$\downarrow$} \\
\midrule

\multirow{3}{*}{\makecell{Qwen-3 (4B)\\+ \copit}}
& \textit{w/o VHR \& LAT}
& 60.87/54.75 & 0.24 \\
& \textit{w/o VH \& SR}
& 50.94/43.40 & 0.28 \\
& \textit{w/o LAT \& SR}
& 44.57/37.20 & 0.33 \\
\midrule

\multirow{3}{*}{\makecell{Ministral-3 (3B)\\+ \copit}}
& \textit{w/o VHR \& LAT}
& 70.09/65.14 & 0.18 \\
& \textit{w/o VHR \& SR}
& 51.91/44.26 & 0.27 \\
& \textit{w/o LAT \& SR}
& 49.33/41.96 & 0.29 \\
\bottomrule
\end{tabular}
}

\caption{Ablation study of pairwise components for Traditional-to-Cyrillic script pivoting, where \textit{VHR}, \textit{LAN}, and \textit{SR} denote Vowel Harmony Recovery, Latin-Assisted Normalization, and Self-Reflection, respectively.}

\label{tab:t2c_pairwise_ablation}
\end{table}

\noindent
\textbf{Traditional-to-Cyrillic Stage Ablation Study.} This analysis isolates the Traditional-to-Cyrillic pivoting stage. Tables~\ref{tab:t2c_single_ablation} and~\ref{tab:t2c_pairwise_ablation} present single- and pairwise-component ablations, evaluated using chrF/chrF++ and Character Error Rate (CER).

Across both backbones, jointly removing Latin-Assisted Normalization and Self-Reflection results in the worst performance, yielding the lowest chrF/chrF++ scores and the highest CER. This suggests that these two components provide complementary contributions to pivoting performance. In contrast, removing Vowel Harmony Recovery together with Latin-Assisted Normalization leads to comparatively better scores, suggesting that Latin-Assisted Normalization provides limited additional benefit when phonological regularization from Vowel Harmony Recovery is already absent.

Compared with single-component ablations in Table \ref{tab:t2c_single_ablation}, pairwise removals in Table \ref{tab:t2c_pairwise_ablation} consistently exhibit substantially lower performance, confirming that Traditional-to-Cyrillic pivoting quality depends on multiple components operating jointly rather than independently.

\begin{table*}[t]
\centering
\begin{tabular}{lccc}
\toprule
\textbf{Language} & \textbf{Artifact Leakage (\%)} & \textbf{Sentence-mode Drift (\%)} & \textbf{Code Mixing (\%)} \\
\midrule
English  & 2.74 & 1.02 & 0.39 \\
Korean   & 0.17 & 0.54 & 2.91 \\
Russian  & 1.19 & 0.37 & 0.37 \\
\midrule
\textbf{Total (N = 24,102)}    & \textbf{4.10} & \textbf{1.93} & \textbf{3.67} \\
\bottomrule
\end{tabular}

\caption{Error rates (\%) across translation outputs in the \copit{}-generated synthetic dataset.}
\label{tab:error_rates}
\end{table*}

\subsection{Assessment of Synthetic Dataset Noisiness}\label{subsec:appendix_errors}
To assess the noisiness of the synthetic dataset of N=8,034 (Section~\ref{subsec:4_2_results}) , we conducted a targeted inspection as an indirect measure of data quality, complementing the downstream performance improvements observed in our experiments. Specifically, we examined three interpretable failure modes: \textit{prompt artifact leakage, sentence-mode drift}, and \textit{code-mixing} across three target languages.

Table~\ref{tab:error_rates} reports the observed error rates. Overall, artifact leakage occurs most frequently across languages (4.10\%), followed by code-mixing (3.67\%) and sentence-mode drift (1.93\%). For Korean, code-mixing appears at a noticeably higher rate than the other error types, while for Russian, artifact leakage constitutes the dominant failure mode. For English, artifact leakage accounts for 2.74\% of outputs, followed by sentence-mode drift (1.02\%) and code-mixing (0.39\%).

Despite these issues, the overall error rates remain relatively low, and this analysis suggests that \copit{} and its generated datasets are still useful in low-resource mitigation scenarios despite some noisiness levels.

\subsection{Improving Forward Translation.}\label{subsec:appendix_trft_full}
\begin{figure*}[t]
\centering
\includegraphics[width=\textwidth]{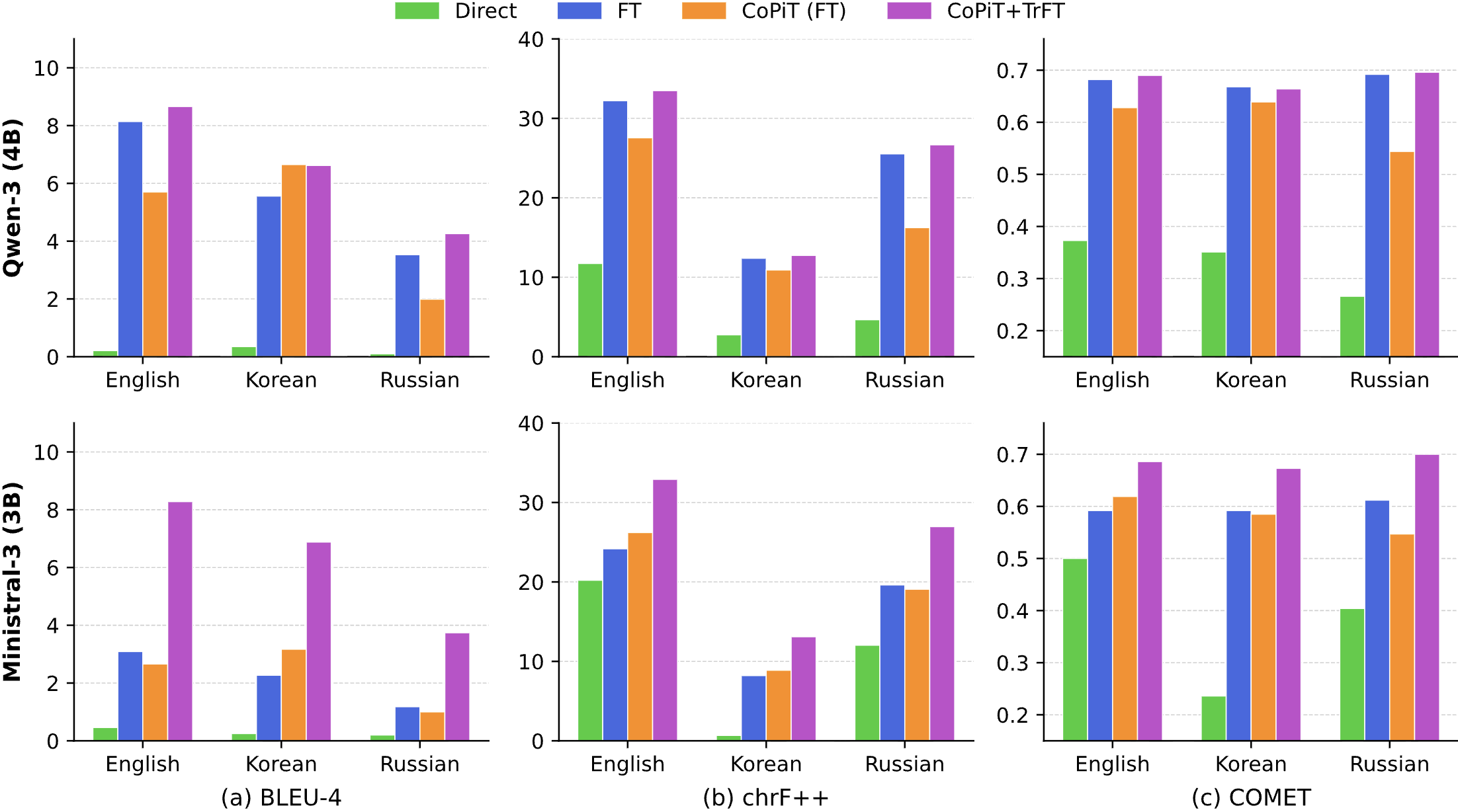}
\caption{Full results across BLEU-4, chrF++, and COMET
under pipeline configurations with and without
translation module fine-tuning, where TrFT denotes
fine-tuning of the translation module on synthetic
parallel data.}
\label{fig:trft_full}
\end{figure*}
Figure~\ref{fig:trft_full} reports full results across BLEU-4, chrF++, and COMET for both backbones. The trends observed in the main paper are largely consistent across all metrics:
CoPiT+TrFT achieves the strongest results in most settings, with particularly pronounced gains in BLEU-4 and chrF++ for English and Korean across both backbones. FT and CoPiT (FT) show mixed relative performance depending on the language and metric; however, we note that this comparison is not data-matched, as FT fine-tunes the translation module directly with \copit{}-generated synthetic data while CoPiT (FT) applies only component-level fine-tuning without translation module fine-tuning. The more informative comparison is between FT and CoPiT+TrFT, which are trained on identical synthetic data but differ in whether the pivot mechanism is applied: CoPiT+TrFT consistently outperforms FT across most languages and metrics, providing cleaner evidence that the pivot mechanism contributes independently of the data.
We note that this comparison operates outside the original low-resource problem setting, in which no Traditional Mongolian--Target parallel data is assumed to be available,
and therefore represents an upper-bound analysis of what becomes possible once the pipeline itself is used to generate synthetic resources.

\subsection{Prompt Templates}\label{subsec:appendix_prompts}
\begin{figure*}[t]
    \centering
    \includegraphics[width=\textwidth]{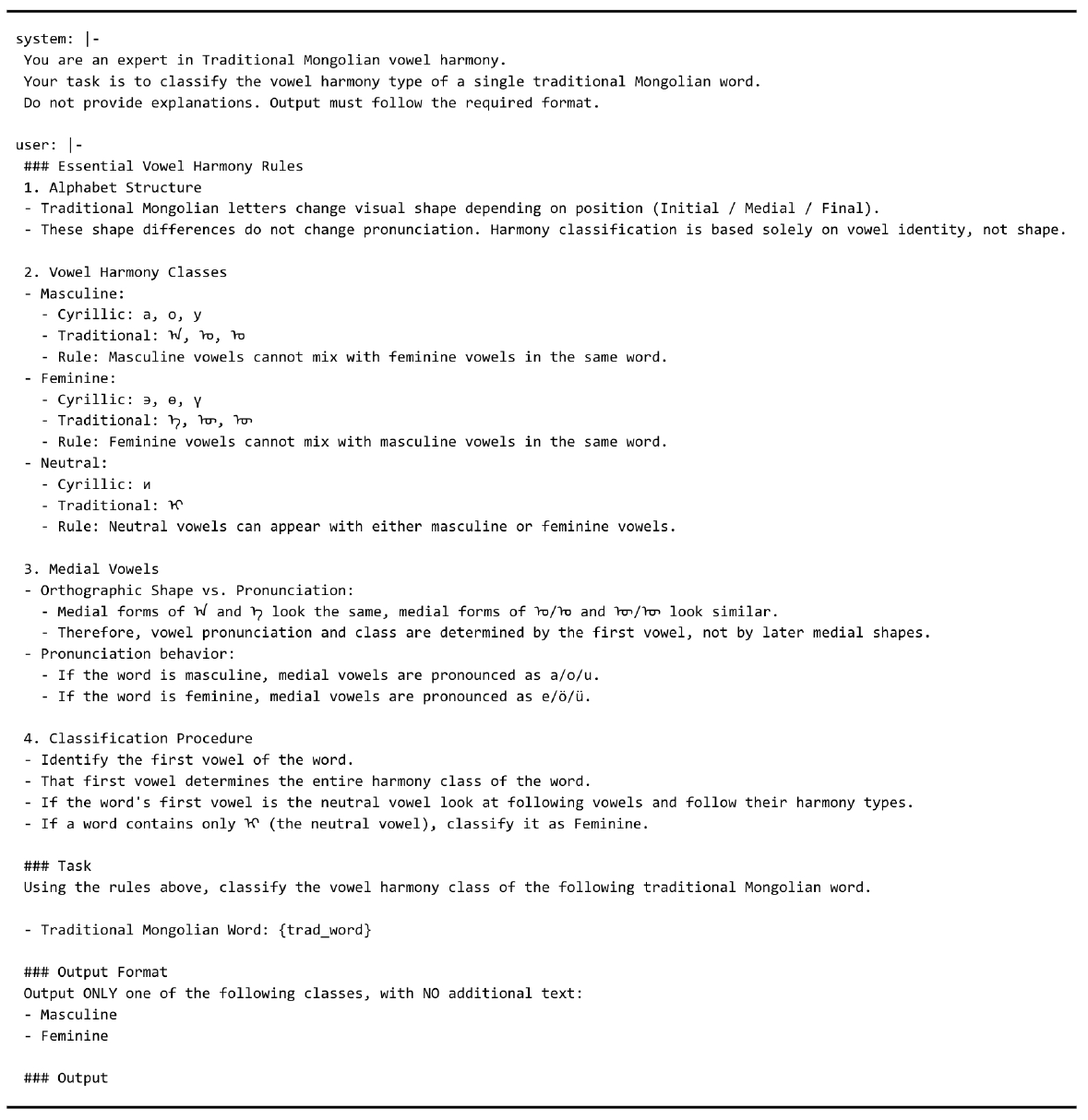}
    \caption{Prompt used for vowel harmony recovery}
    \label{promp:1_vowel_harmony}
\end{figure*}

\begin{figure*}[t]
    \centering
    \includegraphics[width=\textwidth]{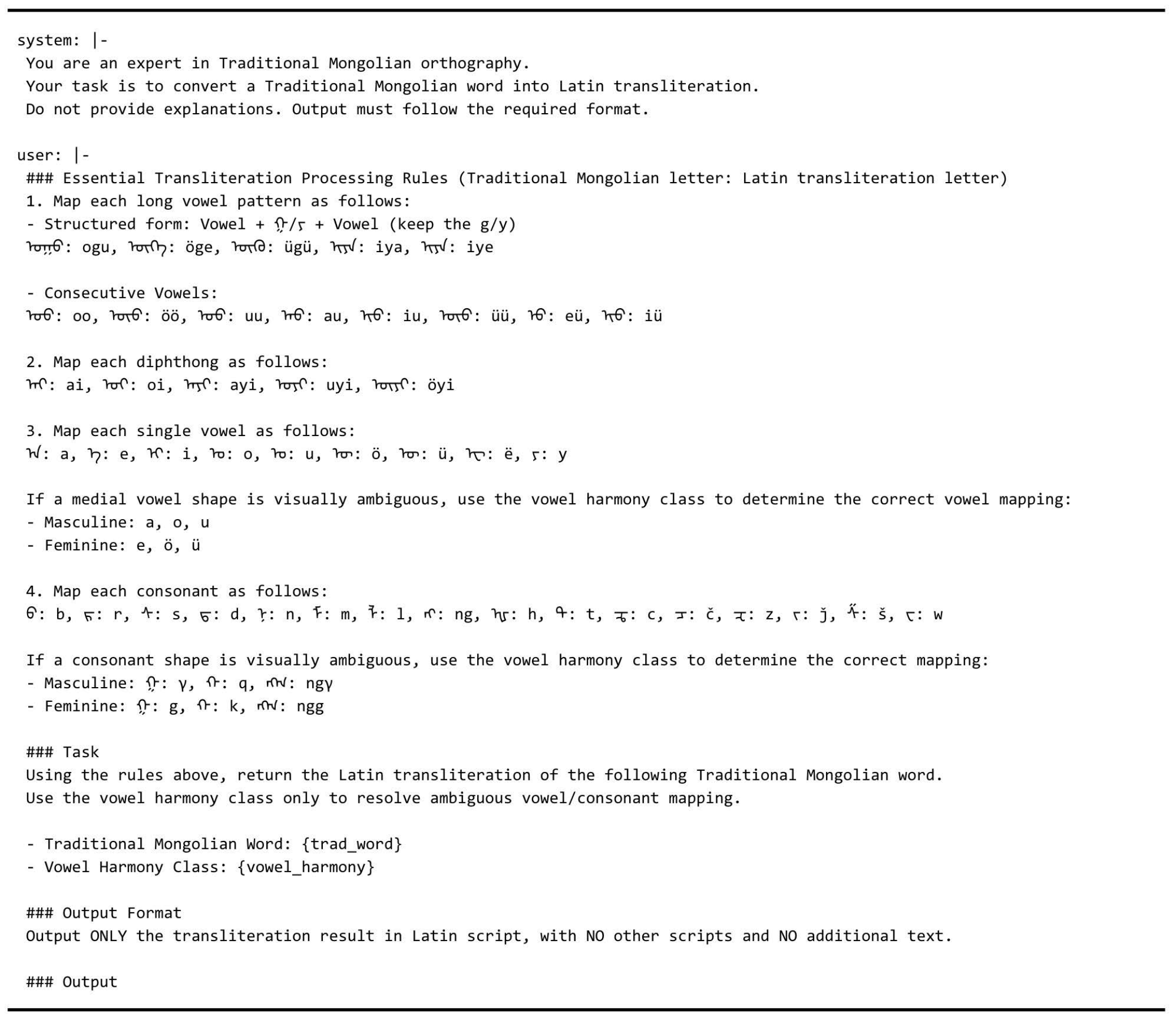}
    \caption{Prompt used for latin normalization}
    \label{promp:2_latin_normalization}
\end{figure*}

\begin{figure*}[t]
    \centering
    \includegraphics[width=\textwidth]{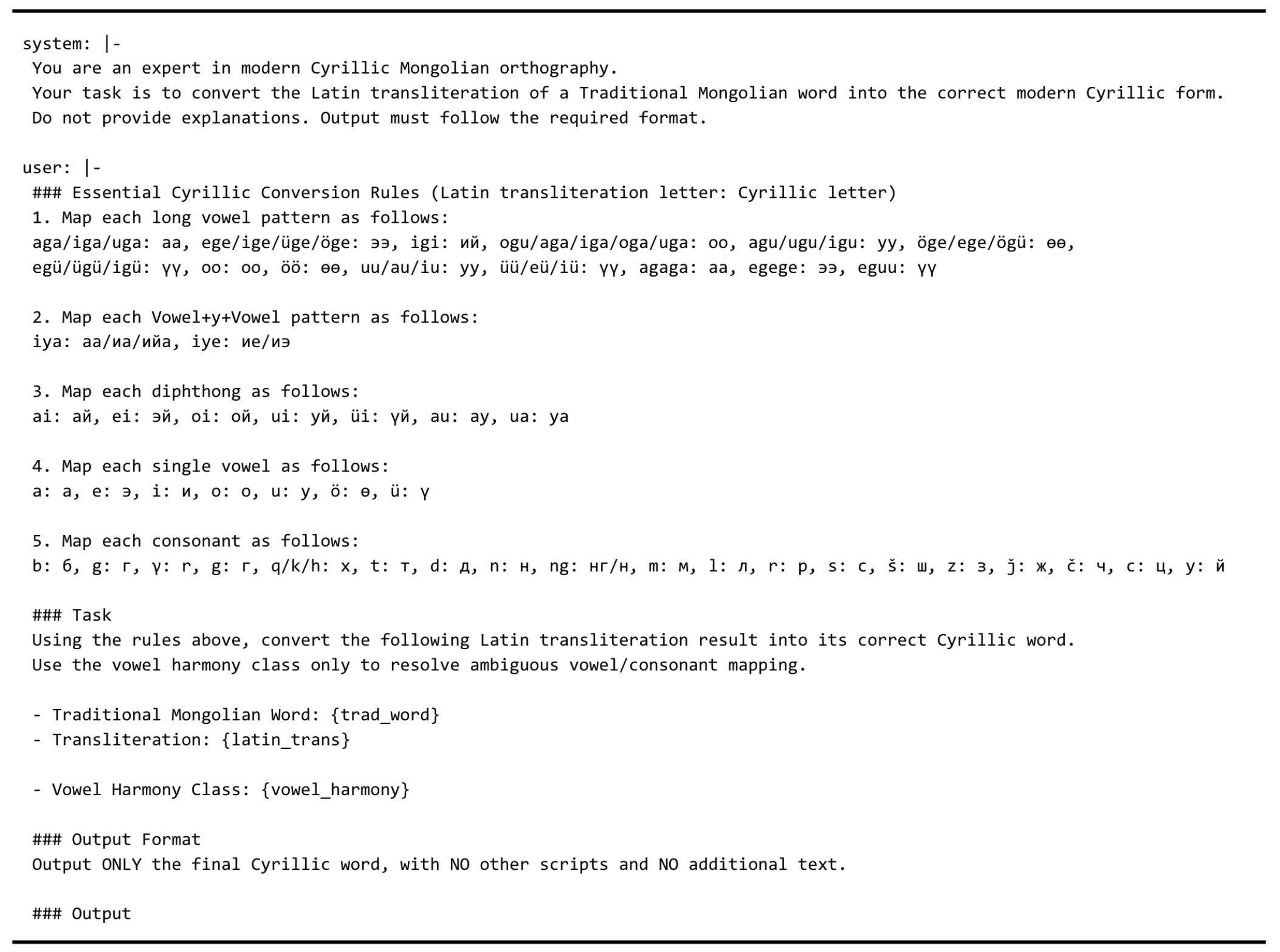}
    \caption{Prompt used for Cyrillic normalization}
    \label{promp:3_cyrillic_normalization}
\end{figure*}

\begin{figure*}[t]
    \centering
    \includegraphics[width=\textwidth]{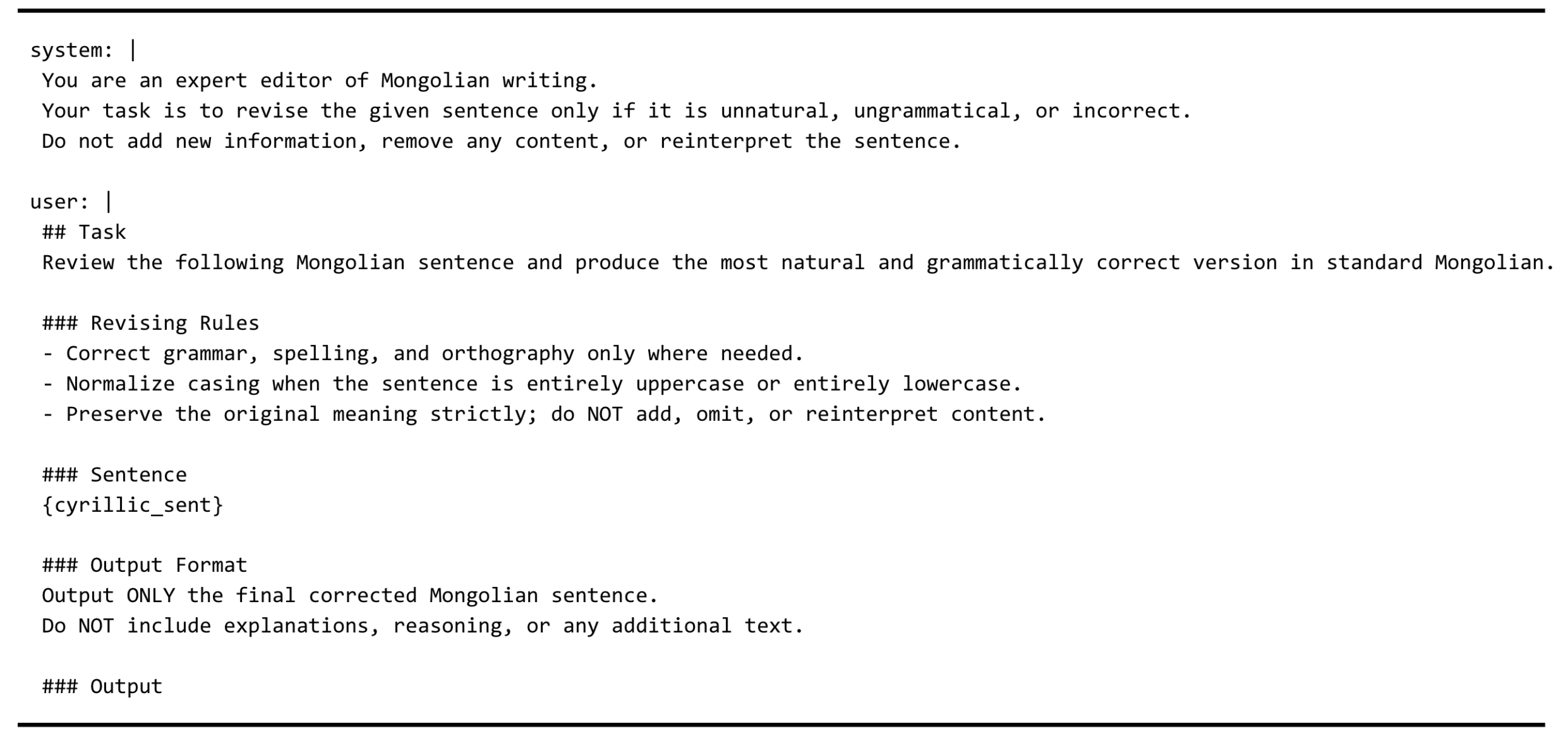}
    \caption{Prompt used for self-reflection}
    \label{promp:4_self_reflection}
\end{figure*}

\begin{figure*}[t]
    \centering
    \includegraphics[width=\textwidth]{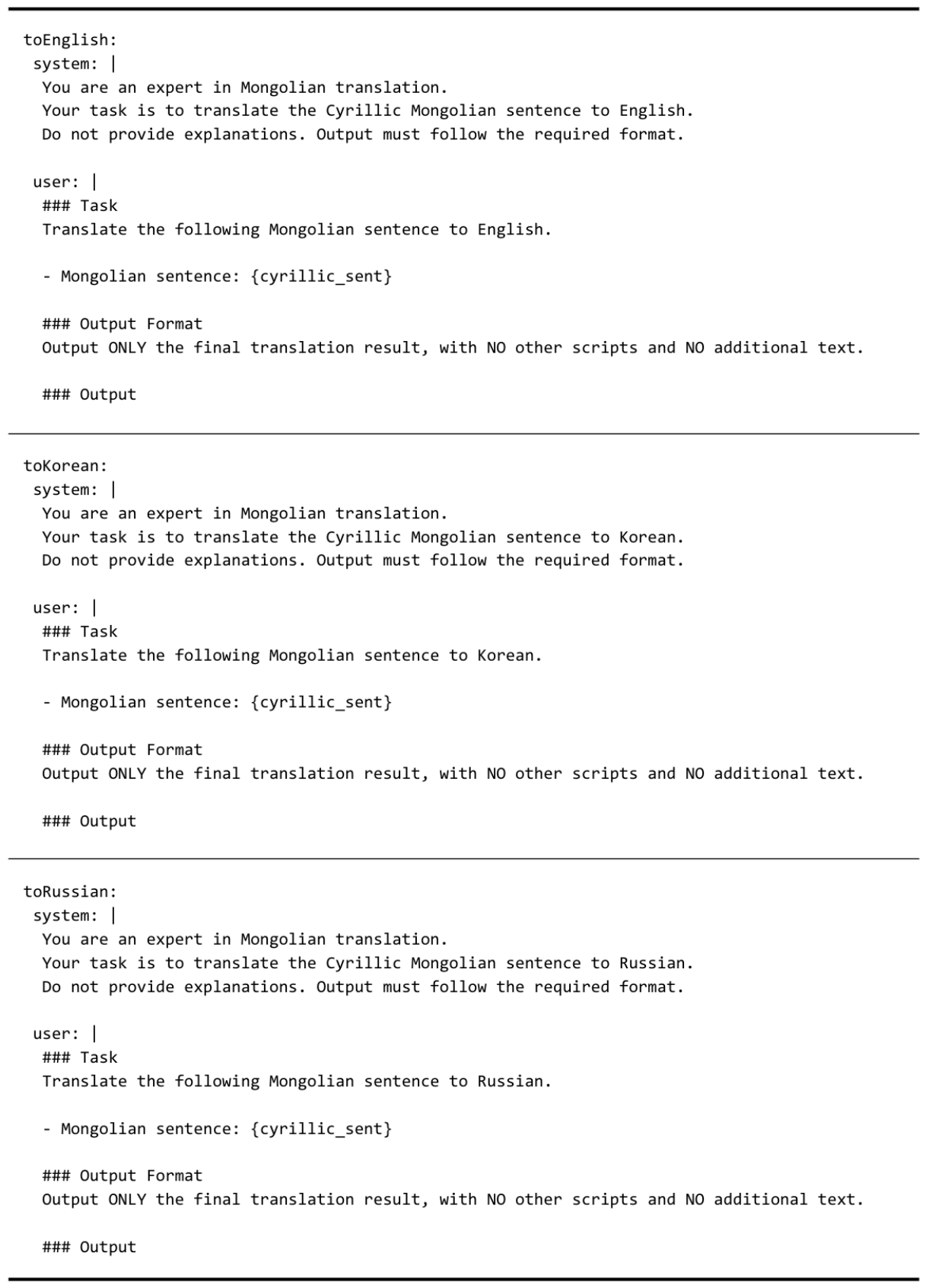}
    \caption{Prompt used for translation}
    \label{promp:5_translation}
\end{figure*}

\begin{figure*}[t]
    \centering
    \includegraphics[width=\textwidth]{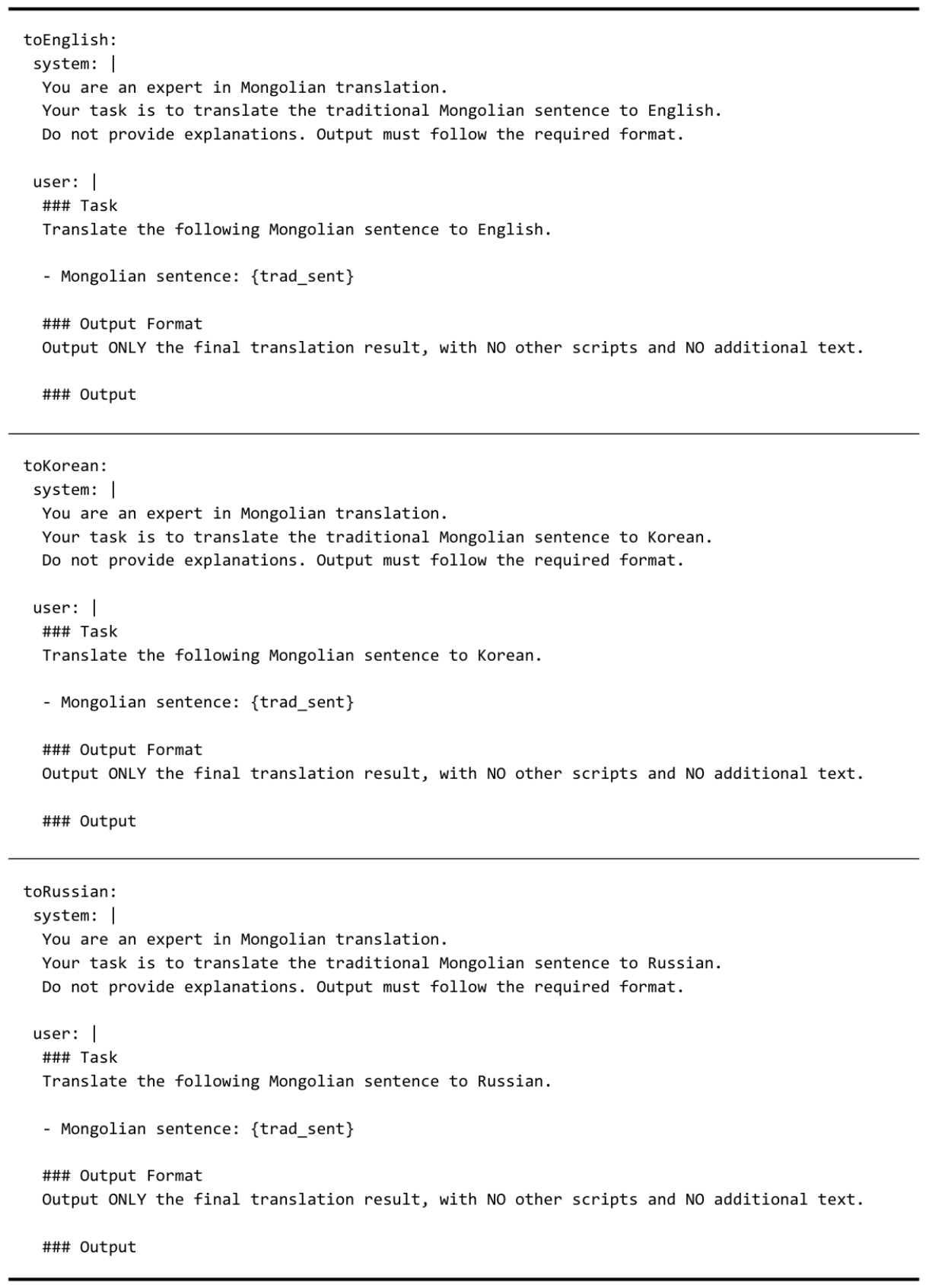}
    \caption{Prompt used for direct translation}
    \label{promp:6_base_translation}
\end{figure*}

This appendix presents the prompts used in our experiments. Figures~\ref{promp:1_vowel_harmony}--\ref{promp:5_translation} show the prompts used in each stage of the \copit{}, while Figure~\ref{promp:6_base_translation} presents the prompt used for the direct translation baseline.

\clearpage

\clearpage
\clearpage
\clearpage
\clearpage
\clearpage

\end{document}